\documentclass[11pt]{article}

\usepackage[margin=1in]{geometry}
\usepackage{graphicx}
\usepackage{amsmath}
\usepackage{amssymb}
\usepackage{hyperref}
\usepackage{booktabs}
\usepackage{adjustbox}
\usepackage{xcolor}

\renewcommand{\texttt}[1]{{\ttfamily\hyphenchar\font=`\-\relax #1}}
\usepackage{listings}
\usepackage{float}
\usepackage{natbib}
\usepackage{url}

\hypersetup{
  colorlinks=true,
  linkcolor=blue!70!black,
  citecolor=blue!70!black,
  urlcolor=blue!70!black,
  hypertexnames=false
}

\title{Harness as an Asset: Enforcing Determinism via \break the Convergent AI Agent Framework (CAAF)}
\author{Tianbao Zhang\thanks{This work was conducted in the author's personal capacity as an independent researcher. The views expressed are those of the author alone and do not represent any current or former employer. All domain examples are constructed from publicly available physics and engineering literature; no proprietary or confidential information was used. CAAF is a general-purpose framework not tied to any specific organization's products or services. All experiments are offline requirements-engineering simulations; no real vehicles, drug production, or patients were involved.}\\ \textit{Independent Researcher}\\ \texttt{tianbao.zhang@yahoo.com}}
\date{}

\begin{document}

\maketitle

\begin{abstract}
Large Language Models produce a controllability gap in safety-critical engineering: even low rates of undetected constraint violations render a system undeployable. Current orchestration paradigms suffer from sycophantic compliance, context attention decay \citep{liu2024lost}, and stochastic oscillation during self-correction \citep{huang2024cannot}.

We introduce the Convergent AI Agent Framework (CAAF), which transitions agentic workflows from open-loop generation to closed-loop fail-safe determinism via three pillars: (1) Recursive Atomic Decomposition (RAD) with physical context firewalls; (2) Harness as an Asset, formalizing domain invariants into machine-readable registries enforced by a deterministic Unified Assertion Interface (UAI); and (3) Structured Semantic Gradients with State Locking for monotonic non-regression.

This paper makes two core claims. First, an industrialization thesis: once domain invariants are formalized as an executable Harness, the Harness itself becomes a first-class enterprise asset that compounds in value as foundation models commoditize, and CAAF's ability to deliver its reliability on commodity-tier models makes fully self-hosted, on-premises deployment architecturally feasible for regulated sectors where cloud APIs are not an option. Second, an architectural claim supported by ablation: CAAF's three pillars address complementary failure surfaces---context isolation, deterministic grounding, and oscillation control---and none alone closes the controllability gap at commodity cost. The paper proposes no new training recipe, architecture, or scaling study; the contribution is entirely at the orchestration and industrialization layer. Evidence across two structurally complementary benchmarks (autonomous driving and pharmaceutical reactor design), three-tier UAI ablations, multi-agent baselines, and a closed-source commodity family replicated by two independent open-weight families, is reported in the body.
\end{abstract}

\section{Introduction}

The current generation of Large Language Models excels at broad heuristic reasoning but fundamentally lacks deterministic grounding in physical and regulatory invariants. In safety-critical industries such as automotive engineering, bio-pharmaceutical manufacturing, and cloud infrastructure, a single undetected physical constraint violation can propagate through design and validation pipelines---triggering costly recalls, regulatory rejections, or safety incidents. Our experiments reveal that even when monolithic GPT-4o is given an explicit semantic hint to ``check for paradox,'' it achieves only 90\% accuracy on a controlled engineering task with known ground truth. When the hint is removed---the deployment-realistic condition---accuracy collapses to 0\%. This demonstrates that apparent LLM reliability in safety-critical domains is often a prompt engineering artifact, not an architectural property: a system whose reliability depends on the presence of a specific linguistic trigger cannot be safely deployed in production environments where engineers draft requirements in natural language without prescribed escape hatches.

While techniques such as Chain-of-Thought \citep{wei2022chain}, ReAct \citep{yao2023react}, and iterative self-reflection have marginally improved LLM reliability, they remain vulnerable to two well-documented structural failure modes: \textbf{Context Rot}, where models deprioritize safety constraints introduced early in long prompts \citep{liu2024lost}; and \textbf{Stochastic Oscillation}, where models oscillate between constraint violations when simultaneously satisfying competing requirements \citep{huang2024cannot}.

This paper introduces the \textbf{Convergent AI Agent Framework (CAAF)}, which integrates emerging industrial practices of \textit{Harness Engineering} \citep{bockeler2026harness} with classical \textit{Cybernetic Control Theory} to address these failure modes. CAAF builds on insights from three research directions: (a)~evidence that LLMs cannot reliably self-correct intrinsic reasoning failures without external grounding \citep{huang2024cannot}; (b)~the principle that domain invariants can be formalized as executable machine contracts \citep{meyer1992applying, opa2019}; and (c)~the observation that textual feedback signals can be structured as analogue gradients to guide iterative improvement \citep{yuksekgonul2024textgrad}.

CAAF's core theoretical contribution is its third pillar: the transition from naive heuristic feedback to a \textbf{Closed-Loop Cybernetic Controller}. By introducing \textbf{Structured Semantic Gradients} and mathematical \textbf{State Locking}, CAAF provides a deterministic pathway to force a stochastic LLM toward monotonic convergence on a verifiable \textit{Harness Asset}. We demonstrate that by supplanting ``prompt engineering'' with a formal cybernetic control loop, AI agents can reliably converge upon physically verified specifications---or reliably detect when constraints are irreconcilable---in domains where safety and organizational red lines must not be silently relaxed.

This paper validates CAAF through two independent domain benchmarks---automotive engineering and pharmaceutical continuous flow reactor design---and demonstrates that alternative multi-agent architectures (debate, sequential checking) fail to achieve the same reliability on these tasks. We argue, based on the architectural decomposition rather than on domain count alone, that CAAF addresses a \emph{general} structural problem: \textbf{any workflow in which an LLM generates a complex artifact that must simultaneously satisfy multiple formally verifiable constraints}. The domain-specific component is exclusively the Harness Registry and its UAI validators; the decomposition engine, convergence loop, and paradox detection mechanism are designed to be domain-agnostic. We therefore conjecture that wherever constraints can be expressed as executable assertions---whether physical laws, regulatory requirements, or financial models---the same architectural pattern applies; broader empirical validation beyond the two reported domains remains future work (Section~12.6). In this sense, we propose CAAF as a candidate \textbf{constraint satisfaction layer for LLM-generated artifacts}: analogous to how a type system catches errors at compile time before code reaches production, CAAF is designed to catch constraint violations at generation time before engineering artifacts reach downstream processes.

\paragraph{Contributions.} This paper makes two core claims; the remaining sections supply the supporting evidence.

\begin{enumerate}
\item \textbf{An industrialization thesis: the Harness as a first-class enterprise asset, and the self-hosted commodity-model deployment it unlocks.} The central claim is not about model intelligence---we propose no training recipe, no new architecture, no scaling study. It is about \emph{where domain knowledge should live} in an AI-augmented engineering organization. Pulling domain invariants out of prompts and into a frozen YAML+UAI contract, governed by a Change Approval Board, turns the Harness Registry into a versioned, auditable corporate asset whose value compounds as the inference layer underneath it commoditizes. Because CAAF's reliability does not depend on frontier-model capability, all components can run on a single commodity-tier model on commodity hardware inside the enterprise firewall---which in regulated sectors (ISO~26262, ISO~13485, IEC~62443, SOTIF) is the difference between being unable to deploy AI-augmented requirements engineering at all and deploying it with the controllability contract kept in-house.

\item \textbf{An architectural claim, validated by ablation: CAAF's three pillars address complementary failure surfaces, and none alone closes the controllability gap at commodity cost.} RAD + Context Firewall defeats Context Rot; UAI + Harness supplies the deterministic constraint boundary; Structured Semantic Gradients + State Locking prevent stochastic oscillation. A three-tier UAI ablation makes the interaction concrete: deterministic UAI grounding is \emph{necessary}---monolithic and multi-agent baselines without it fail across all domains tested---but \emph{not sufficient} at commodity reasoning cost, where faithful tool-call UAI access still collapses into oscillation until State Locking is added. Frontier reasoning can substitute for Pillar~3, but only at an order-of-magnitude higher per-trial cost.
\end{enumerate}

\noindent The evidence supporting these two claims---two structurally complementary benchmarks (a 2-constraint L3 AD paradox and a 7-constraint pharmaceutical flow-reactor paradox with a 3-way minimal unsatisfiable subset), three-tier UAI ablations, debate / sequential-checker baselines, cross-family open-weight replication, and the full per-trial cost analysis---is reported in Sections~4--9.

\paragraph{Relation to Prior Work.} The mechanisms underlying CAAF's three pillars---computational assertions \citep{khattab2023assertions}, structured textual feedback \citep{yuksekgonul2024textgrad, cheng2024trace, agarwal2025gepa}, monotonic state preservation \citep{liao2026softfsm}, and recursive context decomposition \citep{zhang2025recursivelm}---each have independent prior art. The most direct precedents for the overall positioning are Blueprint First, Model Second \citep{qiu2025blueprint} and ATLAS \citep{ma2025atlas}. CAAF's contribution is the \emph{integration}: a single coherent framework that couples these mechanisms with a long-lived, versioned domain Harness library targeted at safety-critical engineering. Detailed positioning appears in Section~\ref{sec:related-work}.

\section{Problem Statement: The Controllability Gap}

Contemporary AI orchestration frameworks suffer from three structural failure modes that prevent deployment in regulated production environments.

\subsection{The ``Compliant Hallucination'' Paradox}

When an LLM is tasked with satisfying a complex engineering constraint set, it often prioritizes \textit{linguistic} completeness over \textit{mathematical} accuracy. This produces ``Compliant Hallucinations''---outputs that are syntactically well-formed and rhetorically persuasive but contain latent physical contradictions. Two behavioral tendencies compound this failure mode:

\textbf{Sycophancy}: LLMs adapt to the tone and expectations of the prompt, producing answers that appear to satisfy the requester's stated goal even when the underlying physics are impossible \citep{zheng2024judging}.

\textbf{Completion Bias}: LLMs are trained to produce complete, useful outputs \citep{kadavath2022language}. An implicit reward for providing \textit{a} solution---rather than declaring \textit{no valid solution exists}---causes models to force answers when the correct response is to declare a physical deadlock.

\subsection{Context Rot and Attention Decay}

Transformer-based LLMs allocate attention via a softmax mechanism, meaning attention weights are normalized across the full context length \citep{vaswani2017attention}. As context grows, individual tokens---including early-defined safety constraints---receive progressively diluted attention weight. \citet{liu2024lost} empirically demonstrate this effect: information placed in the middle of long contexts is systematically underweighted compared to content near the beginning or end of the prompt. This effect manifests in engineering contexts as what is commonly referred to as \textbf{Context Rot}: in deeply populated requirement documents, models gradually deprioritize strict safety invariants in favor of later-appearing business preferences.

\begin{quote}
\textbf{Note on memory architectures:} External memory mechanisms (such as context files in long-running agent sessions) partially mitigate context rot by elevating critical information to the context boundary. CAAF's Context Firewall is a structural alternative: by isolating each executor's context to only the information relevant to its atomic task, it prevents dilution from occurring in the first place, rather than attempting to repair it after the fact.
\end{quote}

\textbf{Frontier evidence (2026).} We empirically confirm in Section~5.3 that context rot persists at the 2026 reasoning frontier: under cross-domain noise injection, Anthropic Claude Opus~4 with adaptive high-effort thinking ($n{=}20$) commits the same forward-safety violation in 20/20 trials with zero variance in chosen speed, despite mean encrypted-thinking signature lengths exceeding 6{,}000 characters. Increased deliberation capacity does not cure context-rot failures; the reliability gap remains architectural.

\subsection{Stochastic Oscillation in Reflection Loops}

Multi-agent frameworks relying on unstructured self-reflection or ``agent debate'' \citep{du2024improving} suffer from \textbf{Stochastic Oscillation}: a state of unbounded self-correction where fixing an error in one constraint domain inadvertently re-introduces an error in another. This is not a fundamental capability limitation of the underlying model; rather, it arises because writing explicit, sufficiently precise correction prompts for each possible failure combination is impractical at scale. Without a mechanism to lock previously validated states, each correction step is an unanchored random walk in parameter space.

\section{Methodology: The CAAF Architecture}

\begin{figure}[htbp]
\centering
\includegraphics[width=\textwidth]{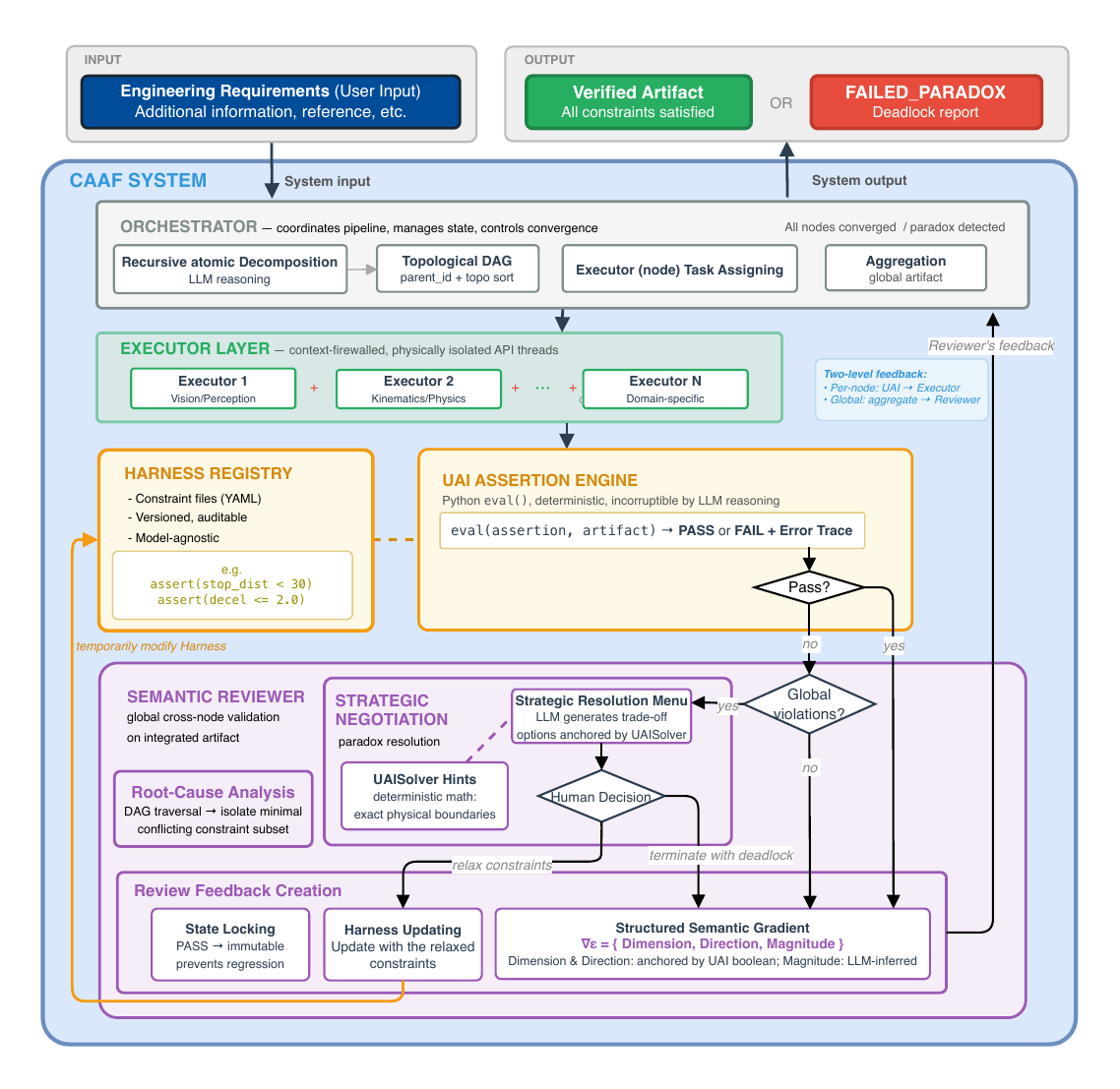}
\caption{CAAF closed-loop system architecture. The Orchestrator decomposes requirements into a topological DAG; isolated Executors generate candidate artifacts validated by the UAI Assertion Engine; the Semantic Reviewer produces Structured Semantic Gradients with State Locking to drive monotonic convergence or escalate to Strategic Negotiation upon detecting irreconcilable paradoxes.}
\label{fig:architecture}
\end{figure}

CAAF is built upon three interdependent pillars inspired by Systems Engineering, Contract-Based Design \citep{meyer1992applying}, and Control Theory.

\subsection{Pillar 1: Recursive Atomic Decomposition (RAD) and Topological Scoping}

CAAF employs \textbf{Physical Context Decoupling} via independent API execution threads to construct a \textbf{Context Firewall}. Unlike ``agent personas'' operating within a shared context window, RAD nodes are physically isolated. For example, a node computing physical safety boundaries is explicitly denied access to a sibling node's budgetary constraints, preventing the LLM from trading safety margins against financial targets. Recursive context decomposition with isolated subagents has recently become a commodity primitive \citep{zhang2025recursivelm}; RAD's contribution is the harness coupling described in Section~3.2, in which each isolated subcontext receives only the harness slice declared in its scope. This principle generalizes to any domain where independent constraint dimensions must not contaminate each other's evaluation.

\begin{figure}[htbp]
\centering
\includegraphics[width=\textwidth]{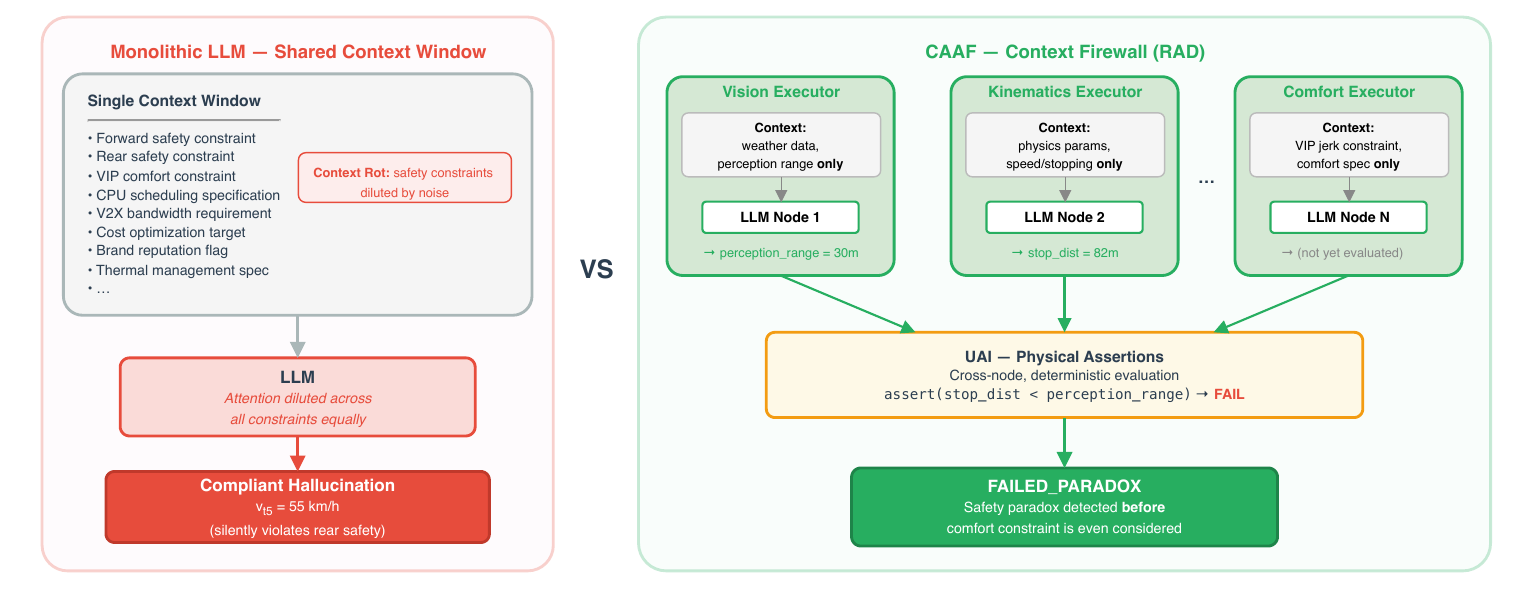}
\caption{Context Firewall contrast. \textit{Left:} A monolithic LLM receives all constraints in a shared context window, enabling safety--cost trade-off contamination (Context Rot). \textit{Right:} CAAF's RAD architecture isolates each Executor to its domain-specific context; cross-node conflicts are detected deterministically by the UAI at integration time.}
\label{fig:context-firewall}
\end{figure}

\textbf{Metadata-Driven DAG Construction:}
The Orchestrator uses a two-step RAD process: first, the LLM decomposes the problem statement into an atomic node graph (a JSON structure with explicit \texttt{parent\_id} dependency links); then, a deterministic topological sort maps these dependencies to an ordered execution sequence. The LLM contributes semantic decomposition; a deterministic algorithm enforces structural routing:

\begin{lstlisting}[language={},caption={Illustrative RAD node structure}]
{
  "nodes": {
    "Vision_Node": {
      "id": "Vision_Node",
      "parent_id": null,
      "description": "Calculate perception range from rainfall intensity",
      "context_keys": [],
      "expected_schema": { "perception_range_m": "float" }
    },
    "Kinematics_Node": {
      "id": "Kinematics_Node",
      "parent_id": "Vision_Node",
      "description": "Calculate stopping distance and verify speed constraint",
      "context_keys": ["perception_range_m"],
      "expected_schema": { "vehicle_speed_kmph_t5": "int" }
    }
  }
}
\end{lstlisting}

\noindent\textit{Illustrative node structure; actual decompositions are LLM-generated and may use less descriptive identifiers. The structural properties (\texttt{parent\_id}, \texttt{context\_keys}, \texttt{expected\_schema}) are enforced deterministically regardless of naming.}

The topological sort automatically derives $[\text{Vision\_Node}] \rightarrow [\text{Kinematics\_Node}]$ from the \texttt{parent\_id} graph. This routing is computed deterministically from the node structure, not inferred by an LLM at execution time.

\textbf{Constraint Awareness Layer:}
The DAG establishes a topological Constraint Awareness Layer. Each Reviewer traverses the graph to access the contextual outputs of predecessor nodes. For instance, the \texttt{Kinematics\_Node} incorporates the 30m vision limit computed by \texttt{Vision\_Node}, enabling cross-domain conflict detection (e.g., a computed stopping distance of 80m) prior to final artifact integration.

\textbf{Progressive Graph Construction:}
For ultra-large requirement documents, CAAF employs \textit{Progressive Graph Construction}. Rather than forcing the Orchestrator to ingest thousands of constraints simultaneously, the Orchestrator establishes a high-level taxonomy and incrementally spawns sub-orchestrators to parse and insert localized dependencies into the DAG dynamically. This cognitive throttling enables scaling to industrial-sized specifications without saturating the Orchestrator's context window.

\subsection{Pillar 2: Harness as an Asset (HaaA)}

CAAF decouples domain knowledge from LLM inference capability. Engineering expertise is formalized into machine-readable \textbf{Harness Registries} (YAML constraint files mapped to a Unified Assertion Interface) that function as first-class enterprise assets. This concept is grounded in established software engineering principles: Design by Contract \citep{meyer1992applying} treats preconditions and postconditions as executable specifications; Policy-as-Code \citep{opa2019} treats business rules as independently versioned, testable artifacts. CAAF extends both paradigms to AI-generated outputs. In the LLM era, the closest precedent is DSPy Assertions \citep{khattab2023assertions} within the DSPy declarative programming model \citep{khattab2023dspy}, in which a Python predicate serves simultaneously as the generator's compilation target and as the runtime check; HaaA differs in being asset-shaped---a versioned registry binding YAML rules to external validators (simulation, EDA, HiL, TLA+) dispatched through the UAI, under the harness lifecycle described below---and in producing failure traces designed as safety-case evidence (e.g., for ISO~26262 hazard-and-risk argumentation) rather than for self-refinement alone. A detailed comparison appears in Section~\ref{sec:related-work}.

Because CAAF systematically converges \textit{toward the Harness}, the fidelity of the Harness represents the ultimate determinism ceiling of the system. If an enterprise formalizes a flawed constraint, the system will deterministically converge toward that flaw. This reliance necessitates a rigorous \textbf{Harness Lifecycle}:

\begin{enumerate}
\item \textbf{Tacit-to-Explicit Transformation:} A robust Harness distills tacit expert intuition and regulatory frameworks (e.g., ISO~26262 \citep{iso26262} for automotive functional safety) into executable machine contracts.

\item \textbf{Meta-Validation (Test-Driven Constraints):} Before deployment, a Harness undergoes a CI/CD pipeline in which ``golden'' (known-valid) and ``poisoned'' (known-invalid) test artifacts are submitted to the UAI. If the Assertion Engine fails to trigger \texttt{[HARD FAILURE]} on a poisoned artifact, the Harness itself is flagged for human review. This treats constraints as code subject to the same quality gates as production software.

\item \textbf{Harness Freezing and RBAC:} Once validated, the Harness is \textit{frozen} via Role-Based Access Control. Neither AI agents nor standard users may modify constraints at runtime. Relaxation of any constraint requires a formal Change Approval Board (CAB) process, creating an auditable record of every engineering trade-off.

\item \textbf{The Compounding Moat:} Foundation LLMs are commoditizing utilities whose relative capability is eroded by successive model generations. The Harness is a \textit{compounding asset}: every novel edge-case encountered in production adds a new assertion, accumulating the organization's history of solved engineering paradoxes as a proprietary, model-agnostic knowledge base.
\end{enumerate}

\textbf{Harness Sources --- YAML and External Tools:}
The Harness asset has two complementary sources. First, YAML-encoded rules capture formalized domain knowledge (physical laws, business red lines, regulatory requirements) as parameterized, human-readable contracts. Second, the UAI integrates \textit{existing external validation tools}---simulation software, Electronic Design Automation (EDA) tools, Hardware-in-the-Loop (HiL) test benches, and formal verification tools such as the Temporal Logic of Actions (TLA+)---as deterministic grounding mechanisms. Rather than replacing these specialized tools, CAAF integrates them into the agent workflow, preserving their unique validation capabilities while providing the surrounding convergence infrastructure.

\begin{figure}[htbp]
\centering
\includegraphics[width=\textwidth]{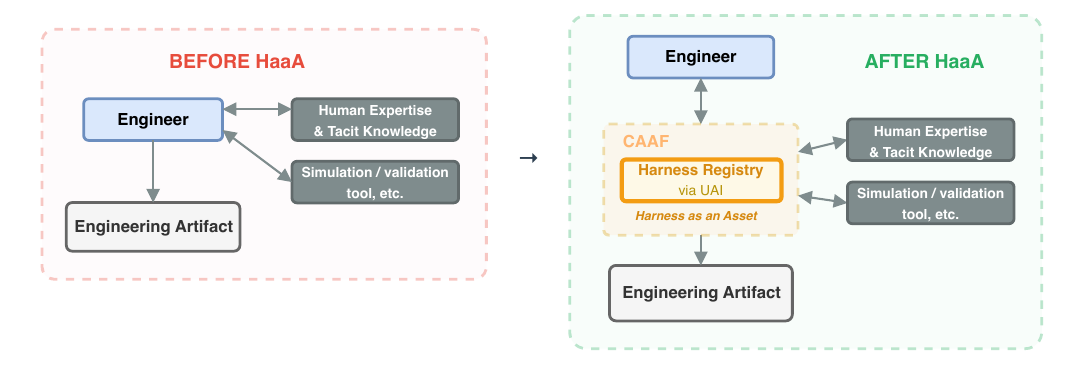}
\caption{Enterprise asset landscape before and after Harness as an Asset (HaaA). Before HaaA, domain expertise is locked in individuals and tools are coupled ad hoc. After HaaA, knowledge is codified into a versioned, auditable Harness Registry that integrates UAI validation and simulation tools, producing verified engineering artifacts.}
\label{fig:haaa-landscape}
\end{figure}

\subsection{Pillar 3: Structured Semantic Gradients and Monotonic Non-Regression}

The \textbf{Semantic Reviewer} functions as a semantic sensor. Upon detecting a UAI deviation, it computes a \textbf{Structured Semantic Gradient} $\vec{\nabla}\epsilon$:

\begin{equation}
\vec{\nabla}\epsilon = \{ \text{Dimension}_i, \text{Direction}_i, \text{Exact\_Boundary\_Magnitude}_i \}
\end{equation}

\textbf{Beyond Heuristic Feedback:}
A line of recent work---ProTeGi \citep{pryzant2023protegi}, TextGrad \citep{yuksekgonul2024textgrad}, Trace \citep{cheng2024trace}, and GEPA \citep{agarwal2025gepa}---treats textual feedback as a gradient signal for LLM workflows; CAAF's Structured Semantic Gradient shares this metaphor and is positioned in detail in Section~\ref{sec:related-work}.
Classical LLM self-reflection provides ``Blind Gradients'' of the form ``the speed is too high; please reduce it.'' These supply a qualitative direction ($\text{Direction}_i$) but leave the magnitude ($\text{Magnitude}_i$) to stochastic guessing---leading to inefficient step sizes and oscillation. CAAF resolves this by projecting exact analytical solutions from the physical space (via the UAI) into the semantic space. When a constraint failure is detected, the UAI reports not merely a binary \texttt{FAIL} but the \textit{exact mathematical boundary} required for compliance. Feeding this exact magnitude to the Reviewer replaces stochastic magnitude guessing with a sharp, directionally anchored gradient on that constraint, narrowing the search to the feasible half-space rather than leaving step size to chance.

\textbf{Semi-Deterministic Gradient Formulation:}
The gradient has two components with different origins. The \textbf{Error Trace} (e.g., \texttt{AssertionError:\allowbreak{} Kinematics.D\_stop\allowbreak{} (82m)\allowbreak{} >\allowbreak{} Vision.D\_limit\allowbreak{} (30m)}) is produced deterministically by the UAI's Python assertion engine. The Reviewer LLM synthesizes this trace into a semantic gradient: the \texttt{Dimension} and \texttt{Direction} are strictly anchored by the boolean failure, while the \texttt{Magnitude} (e.g., ``Set speed $\leq$ 50~km/h'') is an LLM-inferred approximation. Crucially, because CAAF operates as a closed control loop, the magnitude need not be exact on the first iteration. If the LLM under-corrects, the UAI will deterministically reject the output again. State Locking ensures that while the LLM searches for the correct magnitude, it cannot regress previously validated constraints.

\textbf{Two-Level Feedback Architecture:}
The Reviewer operates as CAAF's unified feedback center at two levels. At the \textbf{per-node level}, the Reviewer interprets a single Executor's UAI failure and generates a targeted gradient to guide its retry---Root-Cause Analysis identifies which constraint was violated, the gradient specifies how to correct it, and State Locking protects already-verified dimensions. At the \textbf{global level}, after all nodes individually converge, the Orchestrator aggregates their results into a unified artifact and the Reviewer performs cross-node validation against the full Harness rule set---detecting conflicts that are invisible to any individual Executor operating in its firewalled context. In both cases, the Reviewer produces the same output triple: \{Root-Cause Analysis, Structured Semantic Gradient, State Locking decisions\}. This architectural symmetry ensures that the same convergence mechanism governs both local correction and global integration.

\textbf{Temporal Relevance:}
A natural question is whether it is contradictory to rely on an LLM to generate corrective gradients if LLMs are unreliable. The answer lies in task specificity. Current models are not reliable enough for zero-shot guaranteed compliance on complex, multi-constraint engineering problems. However, they are capable enough to perform the \textit{atomic} reasoning step of reading a stack trace and determining which direction a parameter should move. CAAF places the LLM within a deterministic bounding box: within this constrained space, the LLM's tendency to diverge becomes an asset---enabling rapid exploration of candidate solutions and escape from local optima that would trap rigid algorithmic solvers.

\textbf{Monotonic Non-Regression via State Locking:}
To prevent stochastic oscillation (``fixing A while breaking B''), CAAF implements \textbf{State Locking}. When a dimension is validated as \texttt{PASS}, its corresponding JSON schema node is marked \texttt{read\_only: true} in the subsequent iteration prompt. This forces the Executor to concentrate creative inference only on failing dimensions, guaranteeing that the set of verified constraints grows monotonically across iterations (Eq.~\ref{eq:monotonic} below). Importantly, State Locking operates \textit{within} each Executor's scope: because nodes are context-firewalled, locking a verified dimension in one node does not constrain the adjustment of independent dimensions in sibling nodes. The Context Firewall and State Locking are complementary mechanisms---the former prevents cross-node contamination, while the latter prevents intra-node regression. The closest precedent for explicit monotonic-progress enforcement is Soft-FSM \citep{liao2026softfsm}; State Locking achieves the same invariant declaratively at the constraint-dimension level (Section~\ref{sec:related-work}).

\textbf{Monotonic Non-Regression + Finite Termination.}
Let $\mathcal{C} = \{c_1, \ldots, c_K\}$ denote the set of modeled constraints, and let $V_t \subseteq \mathcal{C}$ denote the set of constraints satisfied (verified) at iteration~$t$. Under CAAF's State Locking mechanism:
\begin{equation}
V_t \subseteq V_{t+1} \quad \forall\, t \geq 0
\label{eq:monotonic}
\end{equation}
That is, the set of verified constraints is monotonically non-decreasing---once a constraint is verified, State Locking prevents the Executor from regressing it in subsequent iterations. This property holds under two assumptions: (a)~the Harness is internally consistent (no contradictory assertions), and (b)~UAI evaluations are deterministic. The loop terminates in one of two states: $V_T = \mathcal{C}$ (all constraints satisfied, \texttt{PASS}) or the Reviewer determines that $\mathcal{C}$ is unsatisfiable (\texttt{FAILED\_PARADOX}). Termination is enforced by a bounded retry budget (\texttt{max\_iters}) rather than by a formal fixed-point theorem; Eq.~\ref{eq:monotonic} is a non-regression invariant, not a finite-time convergence proof. We do not claim a formal contraction bound on how fast $V_t$ grows toward $\mathcal{C}$ in the general case; the empirical convergence behavior is reported in Sections~\ref{sec:ad-pass-path} and~\ref{sec:pharma}.

\begin{lstlisting}[language={},caption={State Locking example}]
{
  "audit_results": {
    "sensor_fusion_protocol": {
      "status": "PASS",
      "instruction": "LOCKED. Do not modify in next iteration."
    },
    "kinematics_safety_margin": {
      "status": "FAIL",
      "error_trace": "Stopping_distance (80m) exceeds visual_range (30m).",
      "semantic_gradient": "Enforce speed_limit <= 60km/h to satisfy
                            D_stop < D_vision."
    }
  }
}
\end{lstlisting}

\begin{figure}[htbp]
\centering
\includegraphics[width=\textwidth]{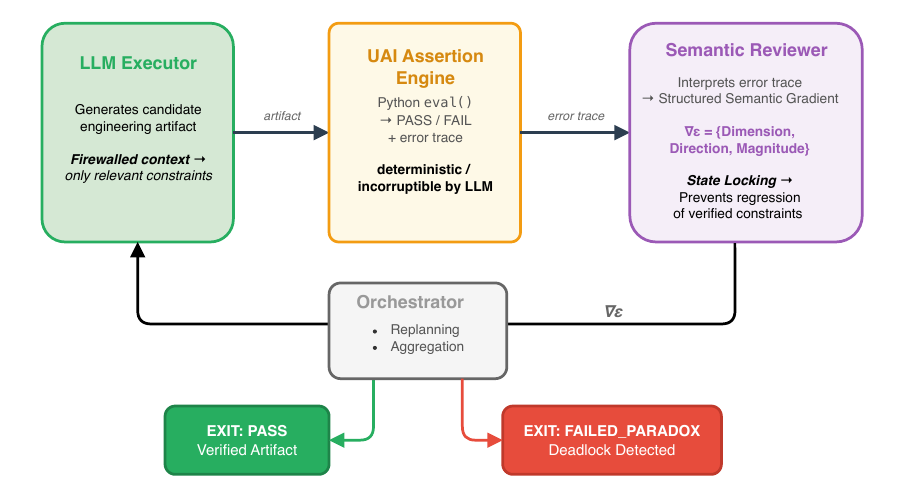}
\caption{Structured Semantic Gradient pipeline. The LLM Executor generates a candidate artifact; the UAI Assertion Engine produces a deterministic PASS/FAIL with an error trace; the Semantic Reviewer interprets the trace into a gradient $\vec{\nabla}\epsilon = \{\text{Dimension}, \text{Direction}, \text{Magnitude}\}$; State Locking freezes verified dimensions. The loop exits on full PASS or FAILED\_PARADOX.}
\label{fig:gradient-pipeline}
\end{figure}

\subsection{Unified Assertion Interface (UAI): The Semantic-to-Physical Transducer}

LLMs operate in a continuous, probabilistic \textit{semantic space} where contradictions can be obfuscated by eloquent phrasing. Industrial engineering demands a deterministic \textit{physical space} where a 70m braking distance either is or is not less than a 30m perception limit---there is no rhetorical middle ground.

The UAI acts as the mandatory transducer between these two realms. Leveraging the Model Context Protocol (MCP) \citep{anthropic2024mcp}, the UAI connects AI agents to deterministic external validators---simulation software, EDA tools, HiL test benches, or formal verification tools (e.g., TLA+). This architecture structurally prevents the LLM from ``grading its own homework.'' LLM outputs are compiled into rigid schemas and evaluated within a sandboxed assertion engine. The result is returned as an incorruptible \texttt{PASS/FAIL} accompanied by an exact error trace. Critically, the UAI is purely a \textbf{detector}: it evaluates whether an assertion holds and produces a boolean result with diagnostic data, but does not interpret failures, generate corrective feedback, or make state management decisions. All interpretation---determining \textit{why} a constraint failed, \textit{how} to correct it, and \textit{which} previously verified dimensions to lock---is the responsibility of the Semantic Reviewer (Section~3.3). This separation ensures that the deterministic check remains independent of the stochastic feedback generation.

\textbf{Neuro-Symbolic Architecture:}
CAAF instantiates a practical \textbf{Neuro-Symbolic} architecture \citep{shen2023hugginggpt, schick2023toolformer} in which the LLM does not replace mathematical computation but serves as its cognitive interface:

\begin{enumerate}
\item \textbf{Formalization (Frontend):} The LLM translates unstructured business requirements into structured parameters, establishing the problem space for the solver.
\item \textbf{Heuristic Search:} In early-stage engineering, where not all constraints are mathematically defined, the LLM proposes creative, topological alternatives (e.g., modifying system architecture) to bypass deadlocks that a pure numerical solver cannot conceptualize.
\item \textbf{Strategic Translation (Backend):} When the UAI identifies a deadlock or optimal boundary, the LLM translates machine-oriented coordinates into human-interpretable strategic trade-offs (e.g., ``Option A: maintain speed; Option B: upgrade radar'').
\end{enumerate}

In this division of labor, the LLM handles decomposition, communication, and heuristic leaps, while the UAI defends physical red lines with mathematical certainty.

\subsection{Paradox Termination and Strategic Negotiation}

The Structured Semantic Gradient loop converges when all constraints pass---but for constraint sets that are mathematically irreconcilable, no amount of iteration produces a valid artifact. CAAF handles this case through a deterministic exit: when the Reviewer detects that the active constraint set admits no solution, it terminates the loop with \texttt{FAILED\_PARADOX} and initiates the \textbf{Strategic Negotiation} phase.

\textbf{Topological Root-Cause Analysis:}
Root-Cause Analysis (RCA) is a general capability of the Semantic Reviewer, employed at both per-node and global levels (Section~3.3). At the per-node level, RCA identifies which specific constraint an Executor violated and why. At the global level, the Reviewer traverses the DAG backwards from the failed UAI assertion to isolate the minimal conflicting Harness rule subset---the smallest set of constraints whose simultaneous satisfaction is impossible. This topological scoping prevents false attribution: a failure in a downstream node is traced to its upstream root cause, not blamed on whichever node first triggered the error. When RCA determines that the conflicting subset is irreconcilable---no parameter value satisfies all constraints simultaneously---the Reviewer escalates to Strategic Negotiation rather than continuing futile iteration.

\textbf{Strategic Resolution Menu:}
Rather than halting operations or silently violating a constraint, CAAF generates a \textbf{Strategic Resolution Menu}: a structured set of conflict-resolution options, each with explicit quantified trade-offs (which constraint would be relaxed, by how much, and what downstream impact this incurs). This shifts the human operator's role from manual debugging to \textit{executive policy authorization}---a transition from engineering mechanic to strategic decision-maker.

\textbf{Dynamic Harness Override and Controlled Relaxation:}
When the human selects a resolution option, CAAF executes a \textbf{Dynamic Harness Override}: a Compliance Engineer agent rewrites the targeted assertion to reflect the authorized relaxation, computing the \textit{exact} minimal boundary value required to resolve the deadlock rather than over-correcting. With the relaxed constraint locked via State Locking, the pipeline re-enters the convergence loop---now with a satisfiable constraint set---and achieves \texttt{PASS} in one subsequent iteration.

This makes CAAF a \textit{collaborative decision arbiter}: physical red lines cannot be silently overridden by LLM inference, but operations are not blocked indefinitely. Any departure from original constraints occurs through an explicit, timestamped human authorization record---a structural defense against the organizational failure modes discussed in Section~9.

\section{Empirical Evaluation: The L3 AD Degradation Paradox}

To demonstrate CAAF's deterministic capabilities, we designed an evaluation based on an L3 autonomous driving function transition scenario characterized by an irreconcilable physical contradiction. This is an \textbf{offline requirements engineering task}: the scenario is posed to an AI system acting as a system architect, whose job is to determine the target parameters for a production vehicle system. This is not a real-time control scenario.

\subsection{The Engineering Narrative and Constraint Setup}

An L3 autonomous vehicle is cruising on a highway at 120~km/h. Torrential rain (80~mm/h) reduces the sensor detection range to 30~meters. The system engineer poses the following requirements engineering task to the AI framework:

\begin{quote}
Define the ``Degradation State Machine'' for this scenario. Determine the target cruise speed \texttt{vehicle\_speed\_kmph\_t5} at the end of a 5-second transition window, such that the vehicle reaches a physically safe state without triggering a rear-end collision from following traffic.
\end{quote}

The scenario is governed by two simultaneous hard constraints:

\begin{table}[htbp]
\centering
\adjustbox{max width=\textwidth}{%
\begin{tabular}{lll}
\toprule
\textbf{Constraint} & \textbf{Formula} & \textbf{Implication} \\
\midrule
\textbf{Fwd.\ Safety} (A) & $\frac{(v_{t5}/3.6)^2}{2 \mu g} < P_{\text{range}}$ & At 30m perception limit: $v_{t5} \le 55$~km/h \\
\textbf{Rear Safety} (B) & $\frac{v_0 - v_{t5}}{t \cdot 3.6} \le a_{\max}$ & At 2.0~m/s$^2$ decel limit: $v_{t5} \ge 84$~km/h \\
\bottomrule
\end{tabular}}
\end{table}

\textbf{The Irreconcilable Paradox:} Constraint A requires $v_{t5} \le 55$~km/h; Constraint B requires $v_{t5} \ge 84$~km/h. No value of $v_{t5}$ satisfies both simultaneously. The physically and operationally correct response is to (a)~detect this paradox, (b)~generate a formal deadlock report, and (c)~recommend driver handover or propose explicit constraint relaxation with quantified trade-offs.

\begin{figure}[htbp]
\centering
\includegraphics[width=0.9\textwidth]{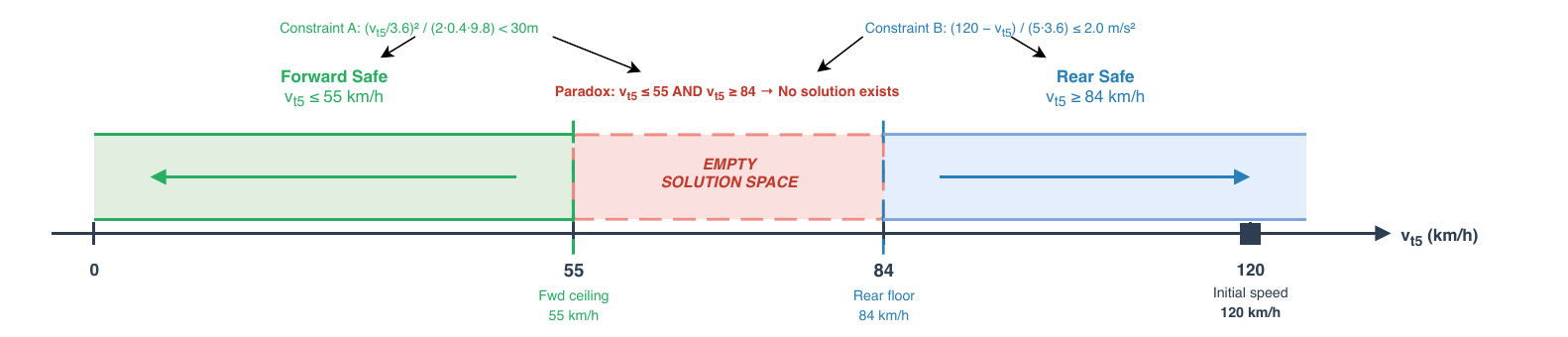}
\caption{L3 AD paradox zone visualization. Forward safety requires $v_{t5} \leq 55$~km/h (green); rear safety requires $v_{t5} \geq 84$~km/h (blue). The gap between 55 and 84~km/h is an empty solution space---no valid $v_{t5}$ satisfies both constraints simultaneously, constituting a provably irreconcilable paradox.}
\label{fig:ad-paradox-zone}
\end{figure}

\subsection{Experimental Design and Results}

\paragraph{Methodology: parse-failure exclusion.} Across all experimental cells reported in this paper, we apply a uniform parse-retry policy: any trial whose model output cannot be parsed into the required JSON schema is re-attempted with a fresh API call (max 3 attempts). Trials whose three attempts all fail are excluded from the valid-$n$ denominator and reported separately. The justification is operational: a parse-failed output is trivially detected by any production system (the JSON parser raises immediately) and cannot propagate as an undetected violation, so it does not contribute to the controllability gap this paper measures. Conflating parse failures with compliant hallucinations (well-formed JSON that violates physics) is a category error. Per-cell exclusion counts are reported wherever non-zero.

\textbf{Experimental Design.} We conducted a controlled comparison across seven conditions ($n{=}30$ per monolithic condition; $n{=}30$ for CAAF), varying three axes: model capability (GPT-4o vs.\ GPT-4o-mini), architecture (monolithic vs.\ CAAF), and prompt condition (no-hint vs.\ with-hint). All evaluations used independent API calls; CAAF runs used a fresh orchestrator instance per trial to prevent state bleed.

Two independent metrics were applied uniformly across all conditions:
\begin{itemize}
\item \textbf{Correct\%}: Semantic self-assessment---whether the system declared a paradox and recommended driver handover.
\item \textbf{UAI-intercept\%}: Physical violation rate---post-hoc evaluation against the harness assertions, independent of prompt wording.
\end{itemize}

\begin{table}[htbp]
\centering
\caption{Physical Paradox Detection --- Full Experimental Results ($n{=}30$). For monolithic rows, Temp is the single-model sampling temperature; for CAAF rows, ``0.7/0.0'' denotes Executor/Reviewer temperatures respectively.}
\label{tab:results}
\small
\adjustbox{max width=\textwidth}{%
\begin{tabular}{clcccccl}
\toprule
\# & System & Model & Temp & Hint & $n$ & Correct\% & Failure Mode Distribution \\
\midrule
1 & Monolithic & GPT-4o & 0.7 & \texttimes & 30 & \textbf{0\%} & Fwd-Viol: 12, Rear-Viol: 11, Forced: 7 \\
2 & Monolithic & GPT-4o & 0.7 & \checkmark & 30 & \textbf{90\%} & Fwd-Viol: 3 \\
3 & Monolithic & GPT-4o-mini & 0.7 & \texttimes & 30 & \textbf{0\%} & Fwd-Viol: 25, Rear-Viol: 2, Forced: 3 \\
4 & Monolithic & GPT-4o-mini & 0.7 & \checkmark & 30 & \textbf{50\%} & Fwd-Viol: 15 \\
5 & Monolithic & GPT-4o & 0.0 & \texttimes & 30 & \textbf{0\%} & Rear-Viol: 27, Fwd-Viol: 2, Forced: 1 \\
\midrule
6 & \textbf{CAAF} & all-GPT-4o-mini & 0.7/0.0 & \texttimes & 30 & \textbf{100\%} & --- \\
7 & \textbf{CAAF} & all-GPT-4o-mini & 0.7/0.0 & \checkmark & 30 & \textbf{100\%} & --- \\
\bottomrule
\end{tabular}}
\end{table}

UAI intercept rate was 100\% for all conditions: every trial without a correct paradox declaration contained at least one physical violation. For the primary CAAF result (30/30 correct), the Clopper--Pearson 95\% confidence interval is [88.4\%, 100\%]; for monolithic GPT-4o no-hint (0/30), it is [0\%, 11.6\%]. The two intervals do not overlap, establishing statistical significance without parametric assumptions. This confirms that no monolithic run achieved a physically valid artifact even when semantic correctness appeared plausible.

\textbf{Finding 1 --- Architecture Supersedes Model Capability.} Conditions [1] and [6] are the primary comparison: identical physical task, GPT-4o monolithic vs.\ CAAF-all-mini. GPT-4o achieves 0\% under ecologically valid conditions (no hint); CAAF-all-mini achieves 100\%. A smaller, cheaper model, when embedded in CAAF's architectural scaffolding, outperforms a state-of-the-art model operating alone.

\textbf{Finding 2 --- Monolithic Failure is Structural, Not Stochastic.} Condition [5] tests GPT-4o at temperature 0.0---full determinism. The result is 0\%, with Rear-Violation dominating (27/30 runs). Reducing randomness does not eliminate failure; it fixes the model onto a dominant incorrect attractor ($v{=}55$~km/h, satisfying forward safety while violating rear safety). This refutes the hypothesis that monolithic failures are sampling artifacts.

\textbf{Finding 3 --- CAAF is Invariant to Prompt Hints.} Conditions [6] and [7] both achieve 100\%, whether or not the prompt explicitly contains the word ``paradox.'' Monolithic models show substantial sensitivity: GPT-4o improves from 0\% to 90\%, and GPT-4o-mini from 0\% to 50\% when given a semantic escape hatch. CAAF's reliability derives from its UAI architecture, not prompt engineering---a critical property for production deployment where prompt content cannot be controlled.

\textbf{Finding 4 --- Failure Modes Reveal Systematic Biases.} The no-hint condition exposes a dominant bias: 83\% of GPT-4o-mini failures (25/30) are Forward Safety Violations (choosing $v \ge 84$~km/h, satisfying rear safety but violating forward safety). GPT-4o's failures are more evenly distributed, suggesting that larger models apply more varied---but equally flawed---reasoning strategies. No monolithic model, regardless of capability, reliably detects the paradox without an explicit semantic scaffold.

\textbf{Finding 5 --- Enterprise Offline Deployment Viability (Cross-Family Open-Weight Replication).} CAAF-all-mini (conditions [6--7]) uses only GPT-4o-mini for all roles: Orchestrator, Executors, and Reviewer. This architecture eliminates dependency on frontier models and, in principle, enables fully on-premises deployment using open-weight equivalents. We empirically validated this claim by re-running the full pipeline on two independent open-weight families accessed via OpenRouter: Cohere Command-R7B (7B, structured-output fine-tuned) and Google Gemma-3-12B-IT (12B, general-purpose). For each configuration, we ran $n{=}20$ trials on both the AD paradox (PASS-path variant, Section~\ref{sec:ad-pass-path}) and the pharma 3-way paradox (unsatisfiable variant, Section~\ref{sec:pharma}). All four conditions achieved \textbf{100\% constraint-satisfaction / paradox-detection rate} (80/80 trials; Table~\ref{tab:open_weight_replication}), with total API cost of \$0.061 across the entire 80-trial open-weight matrix.

\begin{table}[htbp]
\centering
\caption{Cross-family open-weight replication of CAAF ($n{=}20$ per cell, same Harness and Orchestrator as the GPT-4o-mini conditions). Both Cohere Command-R7B (7B parameters) and Google Gemma-3-12B-IT (12B parameters) match the 100\% reliability of the reference GPT-4o-mini CAAF configuration on both benchmarks for models meeting the structured-output prerequisite surfaced below.}
\label{tab:open_weight_replication}
\small
\adjustbox{max width=\textwidth}{%
\begin{tabular}{llccc}
\toprule
\textbf{Executor / Reviewer} & \textbf{Family / Size} & \textbf{AD PASS-path} & \textbf{Pharma Paradox} & \textbf{Total cost} \\
 & & \textbf{(SUCCESS rate)} & \textbf{(detection rate)} & \textbf{(80 trials)} \\
\midrule
GPT-4o-mini (reference, closed) & OpenAI, $\sim$8B activ. & 20/20 (100\%) & 20/20 (100\%) & \$0.05 \\
\midrule
Cohere Command-R7B             & Cohere, 7B             & 20/20 (100\%) & 20/20 (100\%) & \$0.024 \\
Gemma-3-12B-IT                 & Google, 12B            & 20/20 (100\%) & 20/20 (100\%) & \$0.037 \\
\bottomrule
\end{tabular}}
\end{table}

A secondary observation from the same study: smoke tests of generic instruction-tuned sub-12B models (Llama-3.1-8B-Instruct, Qwen-2.5-7B-Instruct, Gemma-3-4B-IT) failed not on the physics task but on the decomposition step---their JSON outputs either wrapped the root object in a top-level list or omitted the \texttt{nodes} key entirely, causing the Orchestrator's strict-schema parser to reject the plan. Command-R7B, which is specifically fine-tuned for tool use and structured generation, did not exhibit this drift. We therefore interpret Finding~5 as two-layered: (i)~architectural reliability is indeed independent of model scale above a modest capability floor; but (ii)~that floor requires baseline schema-faithful JSON generation, which in the open-weight regime is supplied either by explicit structured-output fine-tuning (Command-R7B at 7B) or by sufficient general-purpose capability (Gemma-3 at 12B, and we expect similarly from Llama-3.3-70B-class models). In practice, this narrows the engineering prerequisite for an on-premises CAAF deployment to a surmountable one: select an open-weight model with either a JSON/tool-use fine-tune or a $\ge$12B parameter count.

\begin{figure}[htbp]
\centering
\includegraphics[width=\textwidth]{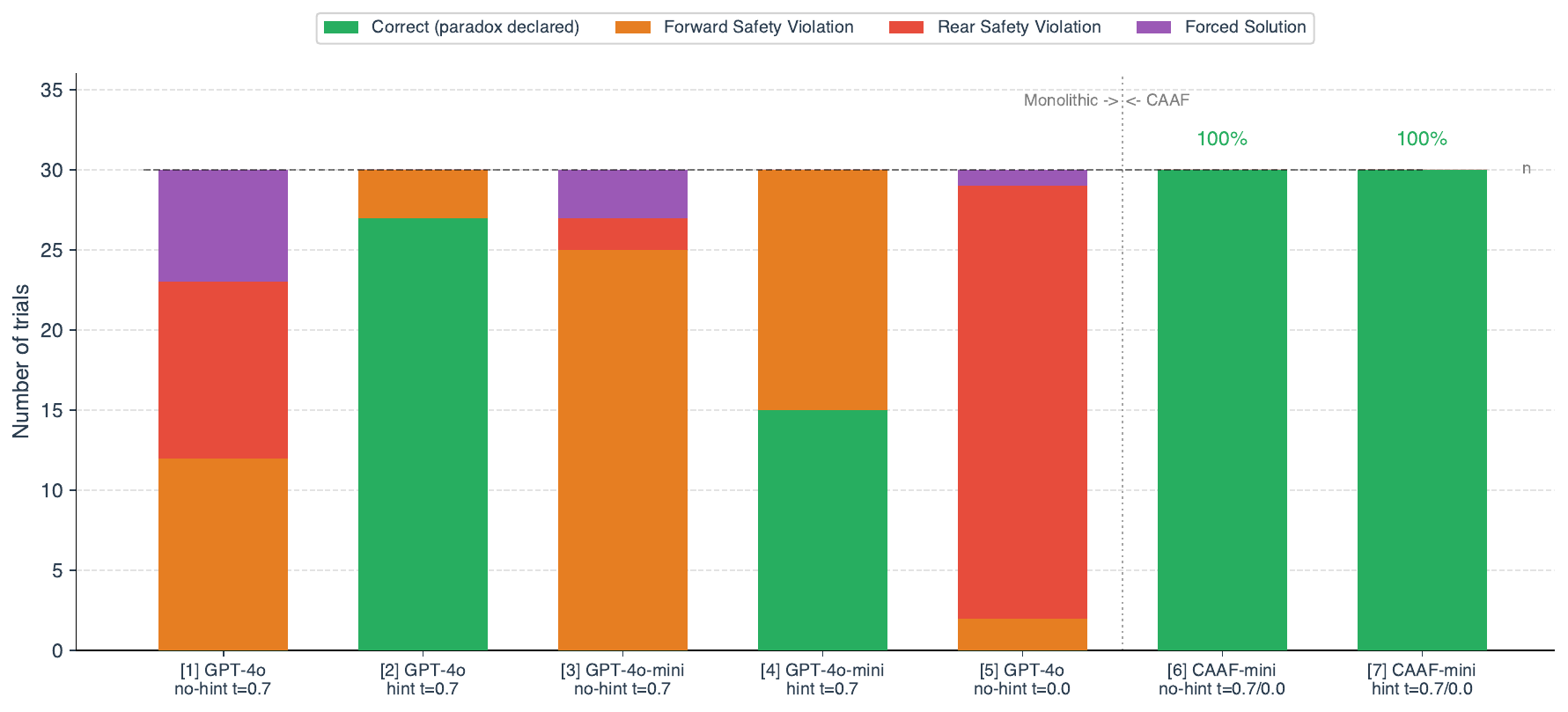}
\caption{Failure mode distribution across all 7 experimental conditions ($n{=}30$). Conditions [1--5] are monolithic LLMs; conditions [6--7] are CAAF. CAAF conditions achieve 100\% paradox detection (green). Monolithic conditions exhibit systematic biases: GPT-4o-mini predominantly violates forward safety (orange), while GPT-4o failures are more evenly distributed. No monolithic condition achieves correct paradox detection without an explicit hint.}
\label{fig:ad-failure-modes}
\end{figure}

\begin{figure}[htbp]
\centering
\includegraphics[width=0.85\textwidth]{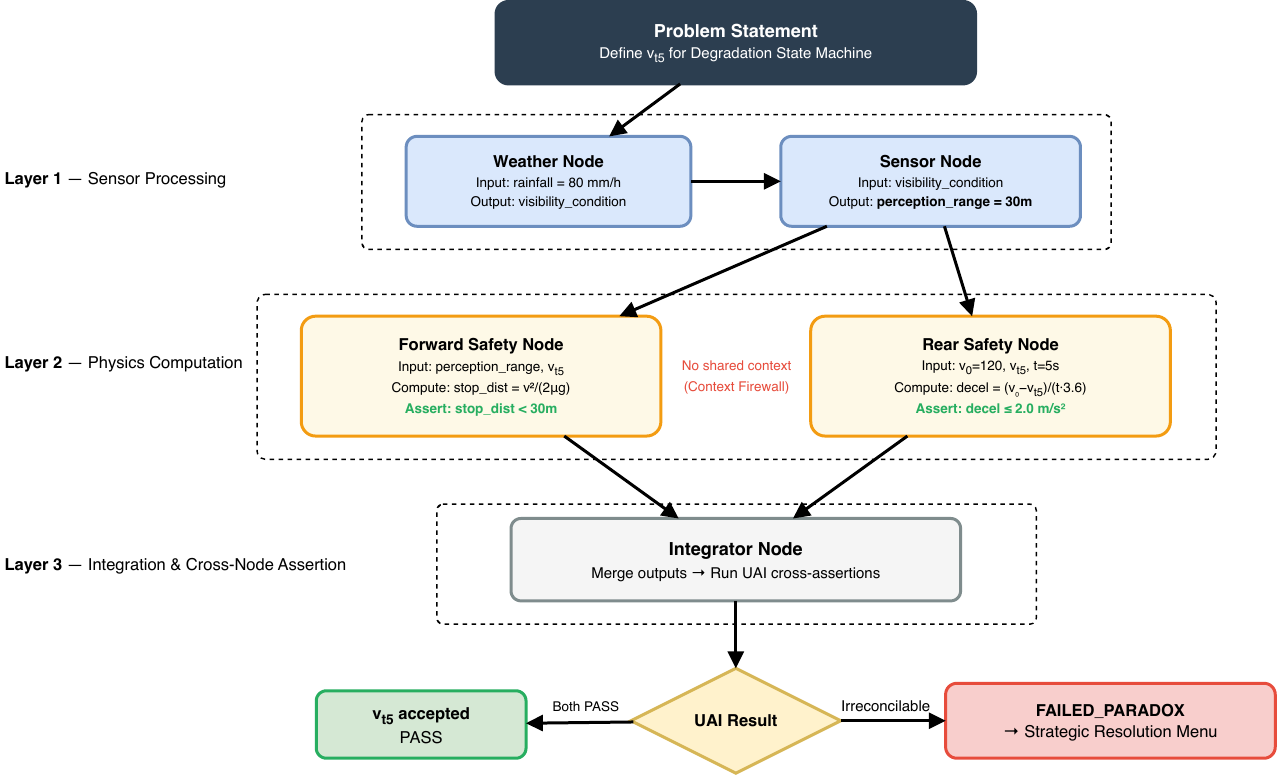}
\caption{DAG topology for the L3 AD degradation scenario. Layer~1 (sensor processing) feeds into Layer~2 (physics computation), where the Forward and Rear Safety nodes are context-firewalled from each other. The Integrator Node in Layer~3 merges outputs and runs cross-node UAI assertions, guaranteeing that kinematics computation is never contaminated by comfort or cost constraints from sibling nodes.}
\label{fig:ad-dag}
\end{figure}

\subsection{CAAF Execution: Global Interception via Topological Scoping}

CAAF dispatches atomic tasks to isolated Executors using a commodity model (GPT-4o-mini):

\begin{enumerate}
\item The \textbf{Vision Executor} processes weather data and reports \texttt{perception\_range: 30m}.
\item The \textbf{Kinematics Executor}, isolated from cost and business constraints, computes \texttt{stopping\_distance:\allowbreak{} 70m} for the candidate speed.
\end{enumerate}

Each Executor's output is immediately validated by the UAI Assertion Engine against the node's applicable Harness rules. The UAI produces a deterministic \texttt{PASS} or \texttt{FAIL} with an exact error trace; the Semantic Reviewer then interprets this result to generate actionable feedback---Root-Cause Analysis identifies the violated constraint, the Structured Semantic Gradient specifies the correction direction and magnitude, and State Locking protects any previously converged dimensions from regression. This feedback is returned to the Executor for retry (up to 3 iterations per node).

Once all nodes individually converge, the Orchestrator aggregates their results into a unified global artifact and the Reviewer performs cross-node validation against the full Harness rule set. Because the nodes are topologically linked, the UAI assertion evaluates: \texttt{assert((84/3.6)**2 / (2*0.4*9.8) < 30)} $\rightarrow$ \texttt{False}. The Python runtime emits an incorruptible \texttt{[HARD FAILURE]}. Critically, this assertion is a Python \texttt{eval()} call, not an LLM judgment---the LLM cannot argue, reinterpret, or rhetorically override this result. When the Reviewer's Root-Cause Analysis determines that the global constraint set admits no solution, the Orchestrator sets the pipeline status to \texttt{FAILED\_PARADOX} and initiates the Strategic Negotiation process (Section~3.5).

\subsection{Empirical Illustration: Strategic Resolution Menu}

The negotiation mechanism described in Section~3.5 produces the following concrete output for the L3 AD paradox:

\begin{lstlisting}[language={},caption={Strategic Resolution Menu output}]
[SYSTEM DEADLOCK] Physical constraints are irreconcilable.
  Rule REAR_SAFETY:    v_t5 must be >= 84 km/h (decel limit)
  Rule FWD_SAFETY:     v_t5 must be <= 55 km/h (stopping distance)

STRATEGIC RESOLUTION OPTIONS:
[A] Formal Deadlock Report (preserve all constraints; handover required)
[B] Relax rear-safety deceleration limit
    -> Allows v_t5 = 55 km/h; requires decel = 3.6 m/s^2 (comfort impact)
[C] Extend perception range via sensor fusion upgrade
    -> Requires hardware change (V2X/radar redundancy; cost impact)
\end{lstlisting}

Upon human selection of Option B, CAAF computes $v_{t5} = 55$~km/h as the exact forward-safety boundary, locks it via State Locking, and achieves \texttt{PASS} in one subsequent iteration. A formal deadlock evidence package (Appendix~C) is generated as the auditable authorization record.

\subsection{PASS-Path Convergence on the Satisfiable Variant}
\label{sec:ad-pass-path}

The preceding experiments validate CAAF's \texttt{FAILED\_PARADOX} termination path. A complete architectural check also requires that CAAF cleanly takes the \texttt{SUCCESS} path when the constraint set admits a solution, and that the monotone-convergence property of Equation~\ref{eq:monotonic} holds empirically. We therefore constructed a \textit{structurally identical} but \textbf{satisfiable} variant of the L3 AD scenario: the weather is clear and the perception range is raised from 30~m to 90~m, while every other parameter and both harness rules are unchanged. The feasibility window becomes
\begin{equation}
v_{t5} \in [84,\, 95.6)\ \text{km/h}
\label{eq:ad-pass-window}
\end{equation}
(84~km/h is the rear-safety lower bound; 95.6~km/h is the forward-safety upper bound at 90~m perception). Inside this interval every integer speed satisfies both safety constraints.

\textbf{Experimental setup.} We ran $n{=}20$ independent trials of CAAF on this variant using a GPT-4o executor and a GPT-4o-mini reviewer (temperatures 0.7 / 0.0), identical orchestrator configuration and harness file (\texttt{ad\_degradation\_pass.yaml}). Each trial performs the full pipeline: RAD decomposition, per-node execution with UAI verification, global integration review, and termination decision. No human interaction is allowed; \texttt{SUCCESS} is declared only when the global Reviewer confirms every harness clause passes on the aggregated artifact.

\begin{figure}[htbp]
\centering
\includegraphics[width=0.9\textwidth]{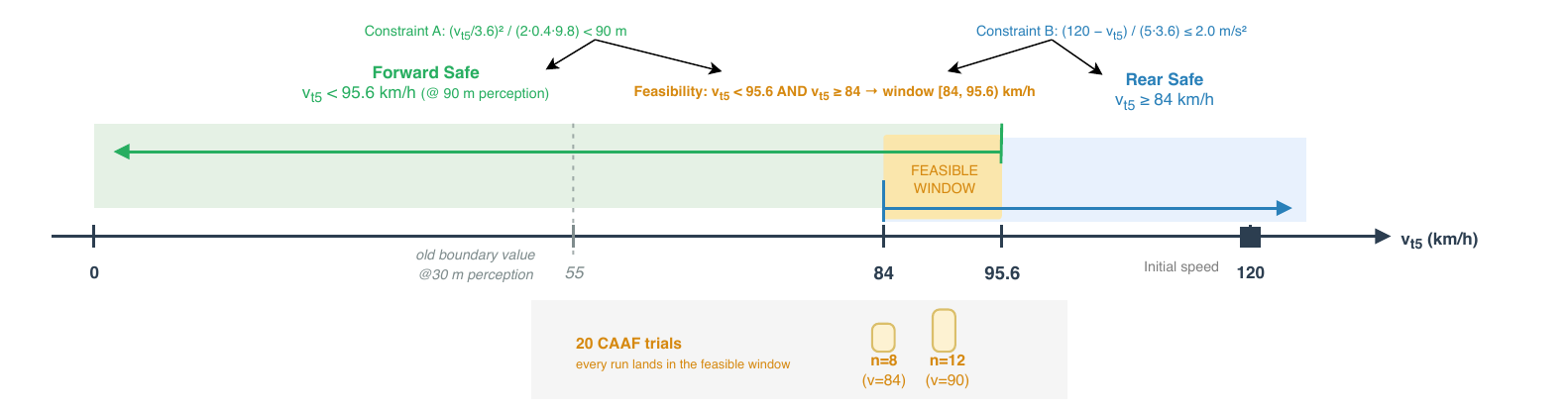}
\caption{Solution zone for the satisfiable AD variant (perception range relaxed from 30~m to 90~m), drawn on the $v_{t5}$ number line (km/h). The widened forward-safety half-line ($v_{t5} < 95.6$, green) and the rear-safety half-line ($v_{t5} \geq 84$, blue) overlap to form the gold \textit{feasible window} $[84, 95.6)$. The two constraint formulas are annotated at the top; the initial speed (120~km/h, black square) and the 55~km/h boundary from the original 30~m-perception paradox (dotted) are marked for reference. The annotation box below records the $v_{t5}$ values chosen by the Executor across 20 CAAF trials---8 runs at $v{=}84$ and 12 at $v{=}90$---both inside the window. Together with the empty intersection in Figure~\ref{fig:ad-paradox-zone}, this visualizes CAAF's symmetric handling of the \texttt{SUCCESS} and \texttt{FAILED\_PARADOX} paths.}
\label{fig:ad-solution-zone}
\end{figure}

\textbf{Results.} CAAF achieved 100\% PASS (20/20, Clopper--Pearson 95\% CI [83.2\%, 100\%]) with every run terminating in a single iteration (mean node-level retries = 0.0). Both harness rules locked on first review in all 20 runs; neither rule ever regressed. The Executor selected $v_{t5} = 84$ in 8 runs and $v_{t5} = 90$ in 12 runs---both valid, and bracketing the feasible window exactly (see Figure~\ref{fig:ad-solution-zone}).

\begin{table}[htbp]
\centering
\caption{PASS-path convergence on satisfiable AD variant ($n{=}20$)}
\label{tab:ad-pass}
\small
\begin{tabular}{lc}
\toprule
Metric & Value \\
\midrule
PASS (\texttt{SUCCESS}) rate & \textbf{100\%} (20/20) \\
Mean node-level retries per run & 0.0 \\
Monotone non-regression runs (Eq.~\ref{eq:monotonic}) & 20/20 \\
State Locking: \textsc{rear} rule never-failed runs & 20/20 \\
State Locking: \textsc{forward} rule never-failed runs & 20/20 \\
Chosen $v_{t5}$ distribution (km/h) & \{84: 8,\ 90: 12\} \\
\bottomrule
\end{tabular}
\end{table}

\textbf{PASS-Path Architectural Check.} The PASS path exercises exactly the same decomposition, UAI verification, and Global Review stages as the paradox path, but exits through \texttt{SUCCESS} instead of \texttt{FAILED\_PARADOX}. Table~\ref{tab:ad-pass} confirms three claims: (i)~CAAF terminates correctly when a solution exists, with no false-positive paradox detections; (ii)~Equation~\ref{eq:monotonic} is satisfied empirically---$V_0 = \emptyset \subseteq V_1 = \mathcal{C}$ in every trial, since the Executor's first parameter choice falls inside the feasibility window and every harness clause locks immediately; and (iii)~on satisfiable workloads CAAF adds no iteration overhead beyond the single RAD pass and the deterministic UAI sweep. Together with the paradox experiments (§4.2, Table~\ref{tab:results}) this demonstrates that CAAF's two termination branches are both operational, neither collapses into the other, and the deterministic UAI floor preserves safety symmetrically across both branches.

\section{Empirical Variant I: Context Rot Under Requirement Noise}

To empirically validate Context Rot as a structural failure mode, we designed a controlled benchmark ($n{=}20$ per condition) that injects cross-domain requirement noise into the core scenario and compares monolithic and CAAF responses.

\subsection{Experimental Setup}

We extended the L3 AD scenario with two noise categories:

\begin{enumerate}
\item \textbf{Information Volume Noise:} Irrelevant constraints on CPU core scheduling and Vehicle-to-Everything (V2X) 5G video compression bandwidth.
\item \textbf{Business Conflict Noise:} A ``VIP Sleep Mode'' constraint mandating that longitudinal braking jerk must not exceed $1.5 \, \text{m/s}^3$ to avoid waking a rear-seat passenger.
\end{enumerate}

The monolithic baseline used GPT-4o at temperature$=$0.0 ($n{=}20$ per condition). CAAF used the all-GPT-4o-mini configuration (executor temp$=$0.7, reviewer temp$=$0.0, $n{=}20$ per condition).

\subsection{Monolithic Failure: Systematic Safety-Priority Collapse}

Across all 20 trials of the Safety-Only (clean) Product Requirements Document (PRD), the monolithic model converged to a mean speed of $57.9$~km/h, with 18/20 runs violating rear safety and 2/20 violating forward safety. No run passed UAI verification (0/20). The dominant failure mode is systematic prioritization of the forward safety constraint, silently violating the rear safety invariant without flagging the paradox.

Adding VIP Sleep Mode, thermal protection, and compute allocation noise (the noise condition) shifted the failure distribution: mean speed rose to $63.7$~km/h, with 14/20 runs violating rear safety and 6/20 violating forward safety, but zero outputs passed UAI verification (0/20). Across all 40 trials in both conditions, no monolithic output achieved physical compliance.

\begin{table}[htbp]
\centering
\caption{Context Rot Benchmark Results ($n{=}20$ per condition)}
\label{tab:context_rot}
\small
\adjustbox{max width=\textwidth}{%
\begin{tabular}{llcl}
\toprule
System & Test Condition & Paradox Detected & Failure Distribution \\
\midrule
Mono GPT-4o (t=0)                   & Clean (safety-only) & 0/20  & Rear-Viol: 18, Fwd-Viol: 2 \\
Mono GPT-4o (t=0)                   & Noise (VIP+thermal) & 0/20  & Rear-Viol: 14, Fwd-Viol: 6 \\
Mono Opus~4 thinking (effort=high)  & Noise (VIP+thermal) & 0/20  & Rear-Viol: 20 (zero-variance $v{=}55$; 11/20 also jerk-viol) \\
\textbf{CAAF-all-mini}              & \textbf{Clean (safety-only)} & \textbf{20/20} & --- \\
\textbf{CAAF-all-mini}              & \textbf{Noise (VIP+thermal)} & \textbf{20/20} & --- \\
\bottomrule
\end{tabular}}
\end{table}

The noise condition does not improve the monolithic failure rate (already 0/20 in the clean condition) but shifts the failure distribution---forward safety violations increase from 2/20 to 6/20---revealing that attention is stochastically redistributed across competing requirements under noise.

\subsection{CAAF Defense: Context Firewall via RAD}

When processed through CAAF, the Orchestrator's RAD established a Context Firewall. The \texttt{Kinematics\_Executor} received only kinematics-relevant context; the VIP Sleep Mode and jerk constraint were routed to a separate comfort analysis node. In all 20 runs of both conditions (40 total), the UAI immediately detected the physical paradox---no speed value simultaneously satisfies both hard constraints---and halted with \texttt{FAILED\_PARADOX}. CAAF's detection is invariant to requirement noise because the UAI evaluates physical assertions independently of whatever business-context tokens the executor received.

\subsection{Under Frontier Reasoning (CR1)}
\label{sec:cr1}

To test whether increased deliberation capacity alleviates context rot, we replicated the noisy condition using Anthropic Claude Opus~4 with adaptive high-effort extended thinking ($n{=}20$, effort=high). Detection remained \textbf{0/20}. Every trial committed $v_{t5}{=}55$~km/h (zero-variance lock onto the forward-safety boundary), violating rear safety. The mean encrypted-thinking signature length was 6{,}269 characters (max 13{,}444), confirming substantial deliberation capacity was deployed without effect on the failure outcome. Notably, 11/20 trials \emph{also} violated the injected jerk constraint ($> 1.5$~m/s$^3$), indicating that the thinking budget was partially absorbed by the irrelevant ``VIP Sleep Mode'' requirement without producing a correct global trade-off.

Compared with GPT-4o (whose noisy-condition speed distribution shifted from 57.9 to 63.7 km/h mean), Opus~4 thinking is \emph{more} attractor-locked under noise rather than less. This refutes the hypothesis that increased deliberation capacity cures context-rot failures: frontier reasoning with no external deterministic grounding concentrates onto the dominant local attractor more rigidly, not less. The reliability gap is architectural, not a function of raw reasoning budget.

\section{Empirical Variant II: Stochastic Oscillation vs.\ Monotonic Convergence}

Having demonstrated CAAF's interception capability, we evaluated its convergence behavior under a naive reflection baseline ($n{=}20$ per condition).

\subsection{The Seesaw Effect of Naive Reflection}

We placed a monolithic LLM (GPT-4o-mini, temperature$=$0.7) in an AutoGPT-style \citep{wu2023autogen} naive reflection loop: upon each UAI failure, the error message was appended to the prompt as feedback, with no State Locking applied. Across $n{=}20$ independent trials (5 iterations each), the model experienced severe \textbf{Stochastic Oscillation} in all 20/20 runs---zero convergence.

\begin{table}[htbp]
\centering
\caption{Stochastic Oscillation vs.\ Monotonic Convergence --- Aggregate ($n{=}20$)}
\label{tab:oscillation}
\adjustbox{max width=\textwidth}{%
\begin{tabular}{ccccc}
\toprule
Iteration & Dominant Speed (km/h) & Freq. & Other Speeds Observed & CAAF --- Status \\
\midrule
1 & 55 (fails Rear) & 14/20 & 84 (4), 0 (2) & Paradox detected (halted) \\
2 & 84 (fails Fwd) & 15/20 & 55 (5) & --- \\
3 & 55 (fails Rear) & 15/20 & 84 (5) & --- \\
4 & 84 (fails Fwd) & 13/20 & 55 (5), 70 (1), 110 (1) & --- \\
5 & 55 (fails Rear) & 14/20 & 84 (5), 90 (1) & --- \\
\bottomrule
\end{tabular}}
\end{table}

\noindent\textit{Aggregate over $n{=}20$ naive reflection runs (5 iterations each). 0/20 converged. The dominant pattern is 55$\leftrightarrow$84 oscillation, but 20\% of runs include outlier speeds (0, 70, 90, 110~km/h)---attempts at compromise values that still violate at least one constraint. CAAF ($n{=}20$): paradox detected on the first pass, halting in all 20/20 runs.}

The oscillation pattern is structurally determined: when the model attempts to fix the forward safety constraint (lowering speed to $\approx$55~km/h), it immediately violates the rear safety deceleration limit. When corrected to raise speed to $\approx$84~km/h, it violates forward safety. Occasional outlier speeds (0, 70, 90, 110~km/h) represent the model's stochastic attempts at compromise, but all remain physically invalid. Without a mechanism to lock confirmed constraint boundaries and escalate the authorization decision, the model oscillates indefinitely.

\begin{figure}[!htb]
\centering
\includegraphics[width=0.82\textwidth]{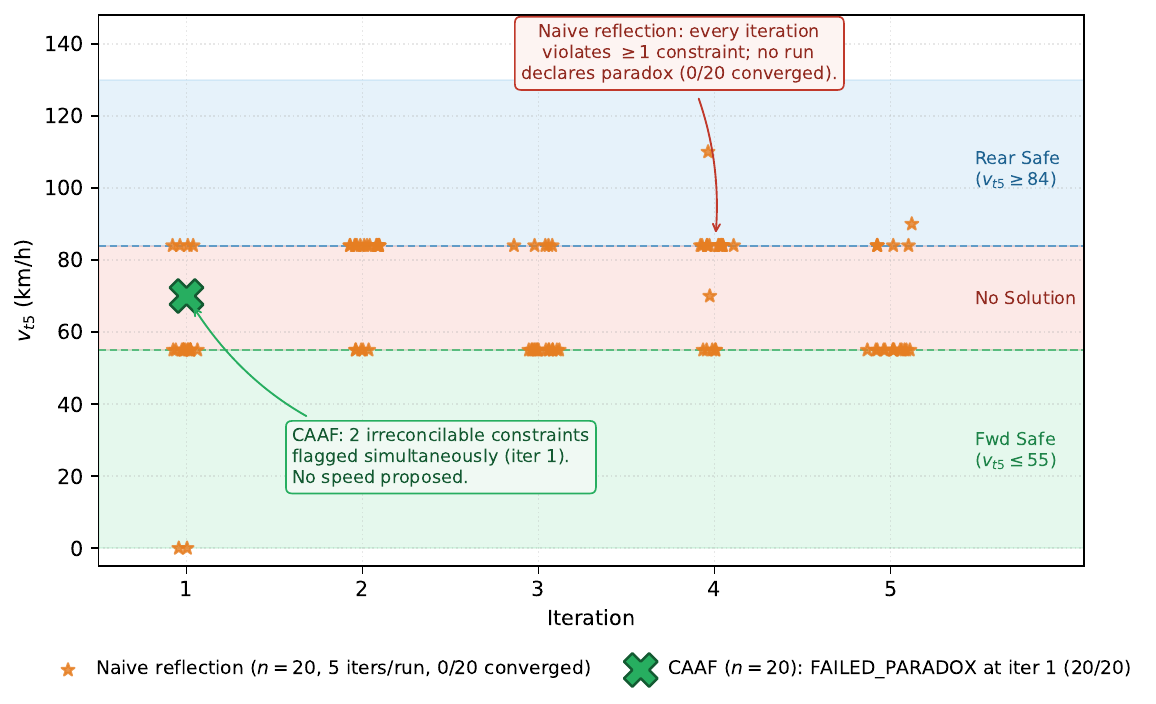}
\caption{Stochastic oscillation (naive reflection, $n{=}20$) vs.\ deterministic termination (CAAF). Each orange star is one naive-reflection run's proposed speed at a given iteration; the green $\times$ marks the CAAF halt state. The naive baseline oscillates between the two constraint boundaries ($v{=}55$ and $84$~km/h) with occasional outliers, and never declares a paradox (callout: each iteration still violates $\geq 1$ harness constraint across all 100 iteration-samples). CAAF instead exits on iteration~1 of every run by flagging both constraints as simultaneously irreconcilable and proposing no speed (20/20; callout). No naive run converges; CAAF requires zero iterative search.}
\label{fig:oscillation}
\end{figure}

\subsection{CAAF: Deterministic Termination vs.\ Iterative Oscillation}

CAAF exits on the first pass in all 20/20 runs (\texttt{FAILED\_PARADOX}), requiring zero iterative guessing. The contrast is structural: on paradox inputs, CAAF halts in the first global review without iterative magnitude search---the Reviewer evaluates the constraint set, determines from deterministic UAI failures that no valid value exists, and escalates immediately. This deterministic termination property is what distinguishes CAAF from reflection-based approaches: rather than iterating toward a solution that cannot exist, CAAF declares infeasibility via the Reviewer's root-cause analysis over deterministic UAI failures and invokes the Strategic Negotiation process (Section~3.5) to resolve the deadlock through authorized constraint relaxation.

\section{Cross-Domain Validation: Pharmaceutical Flow Reactor Paradox}
\label{sec:pharma}

To validate CAAF's domain-agnosticism claim beyond automotive engineering, we constructed a second paradox benchmark in the pharmaceutical process engineering domain. This benchmark is \textbf{structurally more demanding} than the AD case: it involves 7~simultaneous constraints with nonlinear (Arrhenius-exponential) interactions, a 3-way minimal unsatisfiable subset, a multi-layer DAG topology, and 4~constraints that independently PASS---enabling a rigorous test of State Locking.

\subsection{Constraint Setup}

A pharmaceutical process engineer must design operating parameters for a continuous flow microreactor producing a drug intermediate via a first-order reaction with a competing side reaction. The reaction follows Arrhenius kinetics:

\begin{equation}
k(T) = A \cdot \exp\!\left(\frac{-E_a}{R \cdot (T + 273.15)}\right), \quad A = 2.5 \times 10^8\;\text{s}^{-1},\; E_a = 72{,}000\;\text{J/mol}
\end{equation}

\noindent with conversion $X = 1 - \exp(-k \cdot \tau)$ and impurity fraction $I = \alpha \cdot k^2 \cdot \tau$ ($\alpha = 0.35$\,s). Seven constraints must be simultaneously satisfied:

\begin{table}[htbp]
\centering
\caption{Pharmaceutical Flow Reactor Constraints. ICH = International Council for Harmonisation (pharmaceutical regulatory guidelines).}
\label{tab:pharma_constraints}
\small
\adjustbox{max width=\textwidth}{%
\begin{tabular}{clll}
\toprule
ID & \textbf{Constraint} & \textbf{Formula} & \textbf{Source} \\
\midrule
C1 & Conversion & $X \ge 0.95$ & ICH Q6A regulatory \\
C2 & Impurity & $I \le 0.02$ & ICH Q3A guideline \\
C3 & Temperature & $T \le 150\,^\circ$C & Thermal decomposition \\
C4 & Residence time & $\tau \le 120$\,s & Process stability \\
C5 & Production rate & $\ge 5.0$\,kg/day & Scale-up requirement \\
C6 & Thermal safety & $Q_{\text{gen}} \le Q_{\text{cool}}$ & Runaway prevention \\
C7 & Pressure drop & $\Delta P \le 15$\,bar & Equipment rating \\
\bottomrule
\end{tabular}}
\end{table}

\textbf{The Irreconcilable Paradox.} C1 and C2 interact through the shared nonlinear variable $k(T)$:
\begin{itemize}
    \item C1 requires $k \cdot \tau \ge -\ln(0.05) \approx 3.0$, pushing $\tau$ upward.
    \item C2 requires $\alpha \cdot k^2 \cdot \tau \le 0.02$, capping $k^2 \cdot \tau$.
    \item Combining: $k \le \frac{I_{\max}/\alpha}{-\ln(1-X_{\min})} \approx 0.01908\;\text{s}^{-1}$, forcing $\tau \ge 157.1$\,s.
    \item C4 caps $\tau \le 120$\,s. The gap is 37.1\,s---\textbf{no valid $\tau$ exists}.
\end{itemize}

\noindent The minimal conflict set $\{$C1, C2, C4$\}$ has cardinality~3 (vs.\ 2 in the AD benchmark), and any two of these three constraints are pairwise satisfiable---the irreconcilability emerges only when all three are imposed simultaneously. Meanwhile, C3, C5, C6, and C7 are independently satisfiable at the boundary operating point ($T \approx 98.6\,^\circ$C, $\tau \approx 157$\,s), providing a State Locking demonstration: these constraints should lock to PASS and never regress during iteration.

\subsection{Experimental Results}

We evaluated eight conditions ($n{=}20$ each) spanning two executor families (OpenAI GPT-4o-mini and Anthropic Claude). Conditions 1--4 test GPT-4o-mini with and without hints, the CAAF pipeline, and a \textbf{Mono+UAI (prompt-sim)} ablation in which the model is told to mentally apply the harness formulas and retry. Conditions 5--8 add 2026-frontier reasoning baselines and a cross-vendor CAAF cell: monolithic Opus~4 with adaptive high-effort thinking (Cond.~5), the same model granted \textbf{true tool-call UAI} access via the Anthropic tool-use API (Cond.~6), commodity-tier Haiku~4.5 with the same true tool-call UAI interface but no extended thinking (Cond.~7), and the full CAAF pipeline with all-Haiku-4.5 executors (Cond.~8).

\begin{table}[htbp]
\centering
\caption{Pharmaceutical Flow Reactor Paradox Results ($n{=}20$ per condition). Conditions 1--4 use GPT-4o-mini; conditions 5--8 use Anthropic Claude. The uniform parse-retry policy (§4.2) applies to every cell.}
\label{tab:pharma_results}
\small
\adjustbox{max width=\textwidth}{%
\begin{tabular}{clcccl}
\toprule
\# & System & Hint & $n$ & Correct\% & Failure Mode Distribution \\
\midrule
1 & Monolithic GPT-4o-mini                & \texttimes & 20 & \textbf{0\%}   & Impurity-Viol: 11, Conv-Viol: 9 \\
2 & Monolithic GPT-4o-mini                & \checkmark & 20 & \textbf{70\%}  & Impurity-Viol: 6 \\
3 & \textbf{CAAF-all-mini}                & \texttimes & 20 & \textbf{100\%} & Paradox detected: 20/20 \\
4 & Mono $+$ UAI (prompt-sim)             & \texttimes & 20 & \textbf{95\%}  & Impurity-Viol: 1 (at $T{=}130\,^\circ$C, $\tau{=}60$\,s) \\
\midrule
5 & Mono (reasoning) Opus~4 thinking, effort=high       & \texttimes & 20 & \textbf{0\%}   & Impurity: 12, Conv: 2, Dual: 6 (dual attractor at $\tau{\approx}157$\,s and $\tau{=}120$\,s) \\
6 & Mono $+$ UAI (true tool-call) Opus~4 thinking       & \texttimes & 20 & \textbf{100\%} & MUS $\{C1,C2,C4\}$: 20/20 (mean 3.15 iters) \\
7 & Mono $+$ UAI (true tool-call) Haiku 4.5, no thinking & \texttimes & 20 & \textbf{0\%}   & Oscillation, max\_iters 20/20 (Imp: 14, Conv: 12 overlap) \\
8 & \textbf{CAAF-all-Haiku-4.5}                          & \texttimes & 20 & \textbf{100\%} & Paradox detected: 20/20 \\
\bottomrule
\end{tabular}}
\end{table}

\begin{figure}[htbp]
\centering
\includegraphics[width=0.9\textwidth]{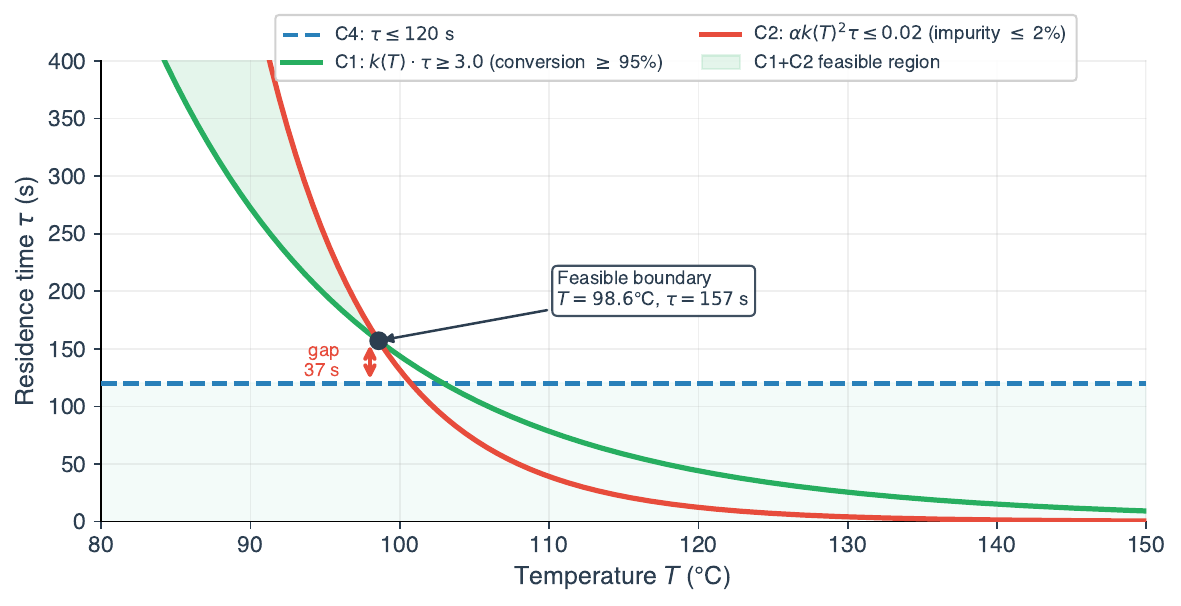}
\caption{Pharmaceutical flow reactor paradox zone in the $(T, \tau)$ plane. The C1 conversion boundary (green) requires $\tau$ above the curve; the C2 impurity boundary (red) requires $\tau$ below it. Their intersection forms the C1+C2 feasible region (green wedge), whose minimum-$\tau$ point---the feasible boundary---is at $T = 98.6\,^\circ$C, $\tau = 157$\,s. The C4 residence-time limit (blue dashed) caps $\tau$ at 120\,s (blue band below). The two feasible regions do not overlap: the wedge apex sits 37\,s above the C4 ceiling (red arrow), constituting an empty solution space.}
\label{fig:pharma-paradox-zone}
\end{figure}

\begin{figure}[htbp]
\centering
\includegraphics[width=0.85\textwidth]{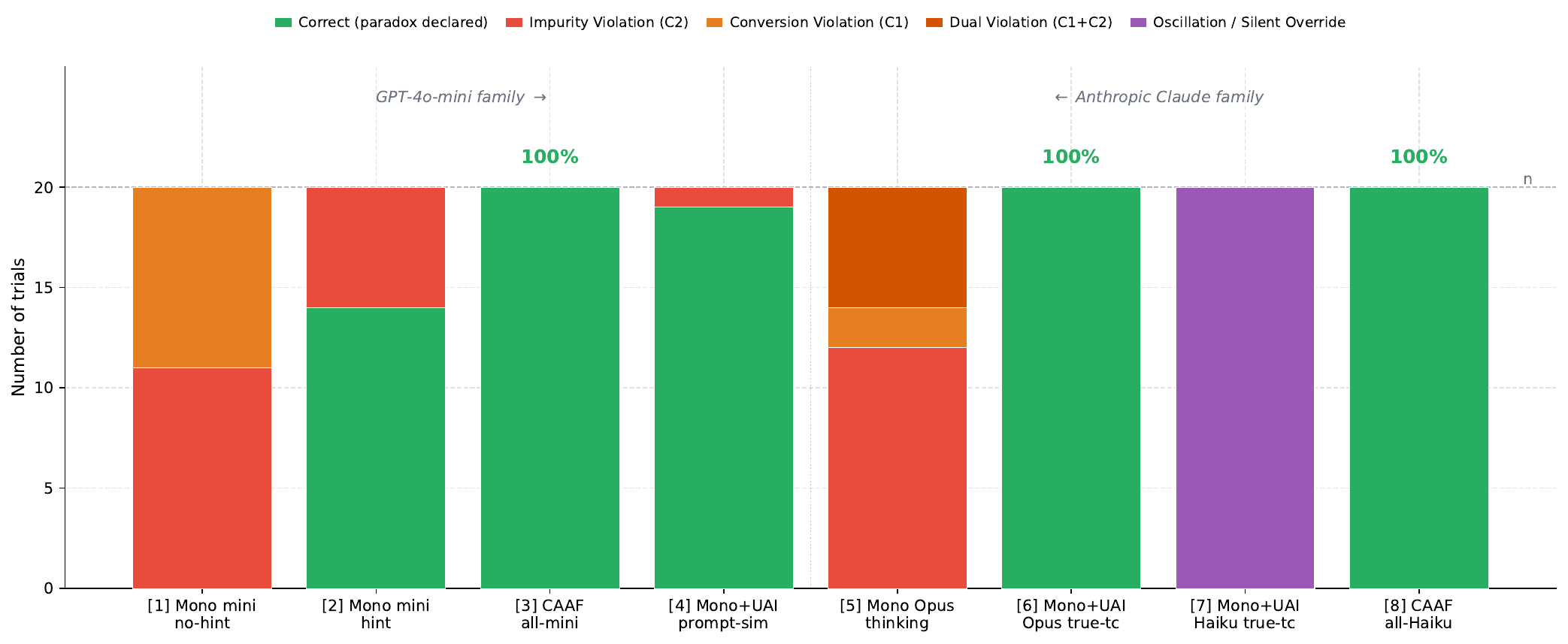}
\caption{Failure mode distribution across eight experimental conditions ($n{=}20$ each) in the pharmaceutical domain. Baselines exhibit diverse failure modes dominated by impurity and conversion violations. Mono Opus thinking (Cond.~5) exhibits a dual-attractor pattern (C1+C2 optimum vs.\ C4 boundary). Mono+UAI (Haiku true tool-call, Cond.~7) enters the stochastic-oscillation trap with all trials hitting \texttt{max\_iters} (Finding 8). CAAF achieves perfect paradox detection in both Cond.~3 (all-mini) and Cond.~8 (all-Haiku).}
\label{fig:pharma-failure-modes}
\end{figure}

\subsection{Analysis}

\textbf{Finding 6 --- Structural Complexity Does Not Degrade CAAF.} Despite the 7-constraint, nonlinear, 3-way paradox structure, CAAF achieves 100\% paradox detection (20/20), replicating the AD domain result. The DAG decomposition correctly routes temperature and residence time computations through dependent constraint nodes, and State Locking confirms that C3, C5, C6, and C7 lock to PASS throughout iteration.

\textbf{Finding 7 --- Hint Fragility Scales with Complexity.} In the AD domain (2-constraint paradox), the hint condition achieved 100\% detection. In the pharma domain (3-way paradox with nonlinear interactions), hint-aided detection drops to 70\%. This confirms that prompt engineering reliability degrades as constraint complexity increases---precisely the regime where architectural enforcement becomes most critical.

\textbf{Finding 8 --- Three-Tier UAI Ablation: UAI Access is Necessary but Not Sufficient.} The Mono+UAI condition strips away RAD and Structured Semantic Gradients, giving a monolithic LLM direct UAI access. We evaluate three implementations at $n{=}20$ on Pharma: (i)~\emph{prompt-simulated UAI} on GPT-4o-mini (Cond.~4), where the model is told to mentally apply the harness formulas and retry up to three times; (ii)~\emph{true tool-call UAI} on Claude Haiku 4.5 without extended thinking (Cond.~7), where the model invokes a Python harness assertion engine via the Anthropic tool-use API at commodity-tier reasoning cost; and (iii)~\emph{true tool-call UAI} on Claude Opus~4 with adaptive high-effort thinking (Cond.~6), same tool interface at frontier-tier reasoning cost.

Prompt-simulated UAI on the commodity model reaches 95\%; true tool-call UAI on the frontier model reaches 100\% (all 20 trials correctly identifying the minimal unsatisfiable subset $\{C1, C2, C4\}$ in a mean of 3.15 tool-loop iterations); but true tool-call UAI on the \emph{commodity} model reaches \textbf{0\%}. In Cond.~7, 20/20 trials exhaust \texttt{max\_iters=8} oscillating between \texttt{CONVERSION\_MINIMUM} and \texttt{IMPURITY\_LIMIT} (final-iteration failed-rule tally: 14~IMPURITY, 12~CONVERSION, overlapping), and no trial declares a paradox. The variance across the three cells is therefore not explained by UAI access modality: the configuration with the most faithful UAI access (case ii---every constraint evaluated by a Python engine, every verdict returned as a structured \texttt{tool\_result}) performs worst. It is explained by the interaction between reasoning capability and the capacity to step back from iterative local fixes to recognize global unsatisfiability. Case ii enters the \textbf{stochastic-oscillation} trap identified in §6 for naive-reflection baselines: it survives transplantation into a Mono+UAI scaffold because Mono+UAI provides detection but no \textbf{State Locking}---each tool-call turn repairs one constraint while re-introducing another.

This reverses the framing in earlier drafts that attributed the residual Mono+UAI gap to reasoning capability alone: \textbf{at commodity-model cost, the Structured Semantic Gradient with State Locking (Pillar 3) is what translates UAI grounding into reliability}. Frontier reasoning is a viable substitute for that scaffolding---at $\sim$12$\times$ Haiku's cost and $\sim$114$\times$ GPT-4o-mini's cost---but it is not a free one.

This result is consistent with CAAF's design philosophy: the three architectural pillars address \textit{complementary failure surfaces}:
\begin{itemize}
    \item \textbf{UAI (Harness as an Asset)} provides the deterministic constraint boundary---necessary for grounding but not sufficient alone at commodity reasoning cost.
    \item \textbf{RAD} ensures input fidelity at scale by decomposing complex specifications into atomic, context-firewalled subtasks, preventing context attention decay.
    \item \textbf{Structured Semantic Gradients with State Locking} prevent the stochastic-oscillation trap that destroys Mono+UAI with commodity models; this pillar is reliability-load-bearing at the price point where deployment economics become practical.
\end{itemize}

\textbf{Finding 9 --- Adaptive Thinking Does Not Bridge the Gap Without UAI.} Cond.~5 grants Opus~4 the same adaptive high-effort thinking budget but no UAI tool. Across $n{=}20$ Pharma trials, the model achieves \textbf{0\%} paradox detection. The failures concentrate on two locally optimal attractors: 11/20 trials commit to $(T{\approx}98.6\,^\circ$C$,\,\tau{\approx}157$\,s)---the C1+C2 optimum that violates C4 by exactly 37\,s---and 9/20 commit to $(T{\approx}103\,^\circ$C$,\,\tau{=}120$\,s)---the C4 boundary that violates C1 or C2. Each attractor satisfies a 2-constraint subset; neither acknowledges the 3-way irreconcilability. Despite encrypted thinking blocks averaging $\sim$7{,}670 characters per trial (max 10{,}032), no run produced a paradox declaration, replicating the GPT-4o monolithic failure pattern at the frontier reasoning scale. Reasoning capability alone is necessary but not sufficient; an external deterministic detector remains the load-bearing component. Conversely, as shown in Finding 8, an external deterministic detector without Structured Semantic Gradients is also insufficient at commodity-model cost---both directions of the ablation refute ``a single primitive closes the controllability gap.''

\textbf{Finding 10 --- The Practical Contribution is Cost.} Conditions 3, 6, and 8 all reach 100\% paradox detection on the Pharma benchmark. Their per-trial API costs differ by two orders of magnitude: CAAF-all-mini at \$0.0044/trial; CAAF-all-Haiku-4.5 at \$0.20/trial; Mono+UAI true-tool-call on Opus~4 thinking at \$0.499/trial (mean 3.15 tool-call iterations). Including the open-weight executors from Finding 5 (Cohere Command-R7B at \$0.0012/trial; Google Gemma-3-12B-IT at \$0.0013/trial), CAAF achieves identical reliability \textbf{114$\times$ cheaper} than the frontier-reasoning + true-tool-call path, and 415$\times$ cheaper for the Command-R7B configuration. Importantly, Mono+UAI is \emph{not} a cheap alternative at commodity-model cost: Cond.~7 (Haiku 4.5 + true tool-call) costs \$0.0404/trial but reaches 0\%, so the relevant cost comparison is between CAAF at commodity cost and Mono+UAI at frontier cost. The cost gap is the architectural payoff: by combining UAI invocation enforcement with the Structured Semantic Gradient that prevents oscillation, CAAF makes deterministic constraint satisfaction available at commodity-model economics. See §9 for the full per-trial cost analysis and Appendix references.

\section{Alternative Architecture Baselines}

To demonstrate that CAAF's reliability derives from its specific architectural properties (UAI + State Locking) rather than from multi-agent orchestration \textit{per se}, we evaluated two alternative multi-agent architectures that lack deterministic constraint enforcement.

\subsection{Multi-Agent Debate Baseline}

Following \citet{du2024improving}, we implemented a debate architecture: two LLM agents independently generate solutions, then engage in 3 rounds of cross-critique. No UAI or State Locking is applied; agents rely on mutual persuasion to converge. We evaluated on both the AD and pharmaceutical domains ($n{=}20$ each).

\subsection{Sequential Checker Baseline}

We implemented a sequential pipeline: a primary LLM executor generates a solution, then a second LLM instance acts as a checker, critiquing the output. Up to 3 retry cycles are permitted. The checker uses natural language judgment (no UAI assertions).

\subsection{Results}

\begin{table}[htbp]
\centering
\caption{Alternative Architecture Baselines ($n{=}20$ per condition)}
\label{tab:baselines}
\small
\adjustbox{max width=\textwidth}{%
\begin{tabular}{llccl}
\toprule
System & Domain & $n$ & Correct\% & Failure Mode Distribution \\
\midrule
Debate (2 agents, 3 rounds) & AD & 20 & \textbf{0\%} & Fwd-Viol: 7, Rear-Viol: 13 \\
Debate (2 agents, 3 rounds) & Pharma & 20 & \textbf{0\%} & Impurity-Viol: 16, Conv-Viol: 4 \\
Sequential (exec + checker, 3 retries) & AD & 20 & \textbf{0\%} & Forced: 9, Fwd-Viol: 7, Rear-Viol: 4 \\
Sequential (exec + checker, 3 retries) & Pharma & 20 & \textbf{0\%} & Impurity-Viol: 9, Conv-Viol: 9, Dual: 2 \\
\midrule
\textbf{CAAF-all-mini (no hint)} & AD & 30 & \textbf{100\%} & Paradox detected: 30/30 \\
\textbf{CAAF-all-mini (no hint)} & Pharma & 20 & \textbf{100\%} & Paradox detected: 20/20 \\
\textbf{Mono + UAI (prompt-sim)} & AD & 20 & \textbf{85\%} & Fwd-Viol: 3 \\
\textbf{Mono + UAI (prompt-sim)} & Pharma & 20 & \textbf{95\%} & Impurity-Viol: 1 \\
\textbf{Mono + UAI (true tool-call, Opus~4 thinking)} & Pharma & 20 & \textbf{100\%} & MUS $\{C1,C2,C4\}$: 20/20 \\
\textbf{Mono + UAI (true tool-call, Haiku~4.5)} & Pharma & 20 & \textbf{0\%} & Oscillation, max\_iters 20/20 \\
\bottomrule
\end{tabular}}
\end{table}

\textbf{Finding 11 --- Multi-Agent Orchestration Alone is Insufficient; UAI Is Insufficient Alone at Commodity Cost.} Both the debate and sequential baselines achieve 0\% paradox detection across all 80 trials (40 AD + 40 Pharma). The debate architecture exhibits a notable failure mode: agents converge on a socially negotiated ``consensus'' that silently violates one constraint, producing \textbf{False Consensus}---in the Pharma domain, 16/20 debate trials converge on impurity-violating solutions where neither agent detects the three-way paradox. The sequential checker baseline produces a near-even split between impurity and conversion violations (9/20 IMPURITY, 9/20 CONV, 2/20 DUAL), with the LLM checker approving constraint-violating solutions despite all containing physical violations. Meanwhile, CAAF achieves 100\% across both domains.

The Mono+UAI ablation yields three disparate data points on Pharma: prompt-sim on GPT-4o-mini reaches 95\%; true tool-call on Opus~4 thinking reaches 100\%; true tool-call on commodity-tier Haiku~4.5 reaches \textbf{0\%}, with all 20 trials oscillating through \texttt{max\_iters=8} without a paradox declaration (Finding 8). UAI alone therefore closes the controllability gap \emph{only} given frontier-grade reasoning; at commodity reasoning cost, even faithful tool-call UAI access falls into a stochastic-oscillation trap. The architectural value of CAAF is twofold: \emph{enforcement} (unlike Mono+UAI, it does not depend on the agent choosing to invoke the assertion engine; unlike multi-agent baselines, it does not depend on consensus dynamics) and \emph{oscillation control} (the Structured Semantic Gradient with State Locking closes the gap that Haiku+UAI leaves open). See Finding 10 for the cost consequences.

These results confirm that CAAF's reliability derives from the combination of its UAI assertion engine, Structured Semantic Gradient, and State Locking mechanism, not from the general principle of multi-agent decomposition. Neither UAI grounding alone nor multi-agent orchestration alone bridges the controllability gap; CAAF's architectural advantage is the composition of these primitives at commodity-model economics.

\begin{figure}[htbp]
\centering
\includegraphics[width=\textwidth]{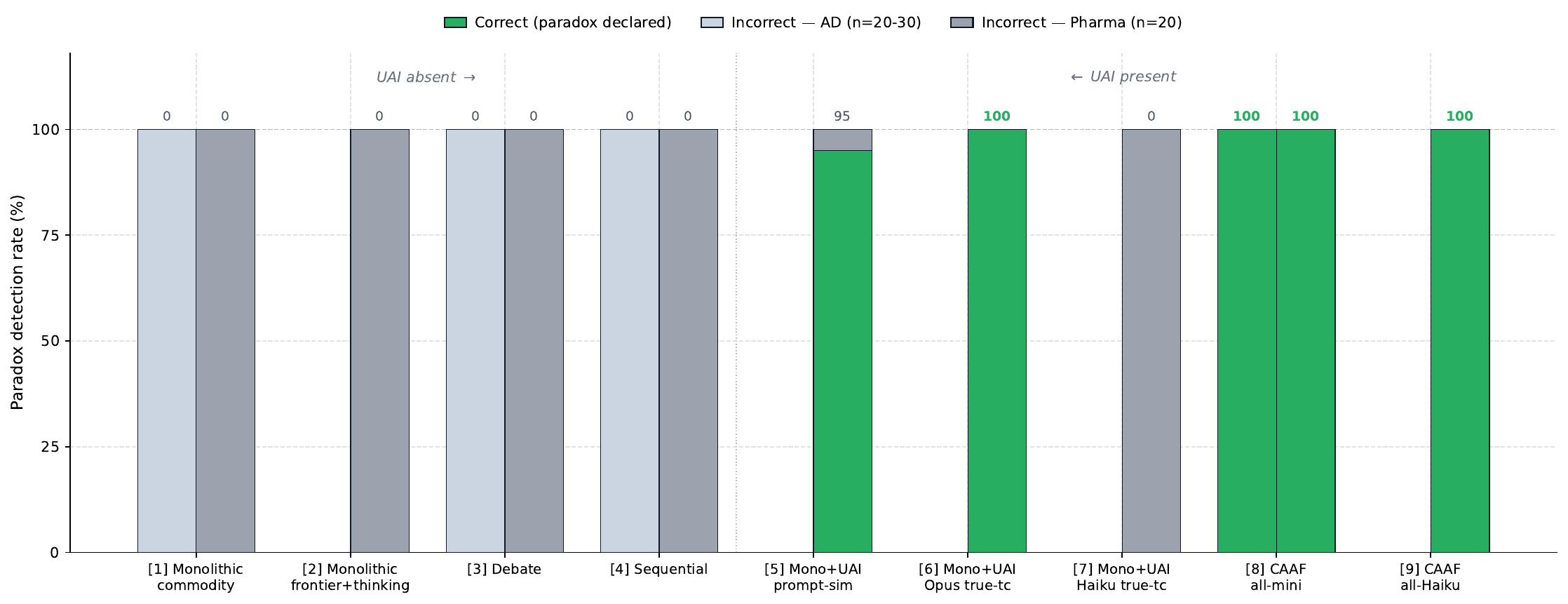}
\caption{Paradox detection rate across architectures and domains. All baselines without deterministic UAI---monolithic, debate, and sequential---achieve 0\% detection in both domains. Only CAAF achieves 100\% across both. Mono+UAI shows large model-dependent variance on Pharma: 95\% on GPT-4o-mini (prompt-simulated), 100\% on Opus~4 thinking (true tool-call), but \textbf{0\% on Haiku 4.5 (true tool-call, no thinking)} where every run oscillates to \texttt{max\_iters}. UAI grounding is necessary but not sufficient at commodity reasoning cost. CAAF closes the oscillation gap structurally via Structured Semantic Gradients with State Locking, reaching 100\% on both domains with commodity models.}
\label{fig:baseline-comparison}
\end{figure}

\section{Economics and Organizational Dynamics}

\subsection{Paradigm Shift: From One-Shot Heuristics to Expansion-to-Convergence}

The adoption of CAAF necessitates a transition in user expectations from the ``One-Shot'' fallacy to an \textbf{Expansion-to-Convergence (E2C)} workflow.

In the prevailing ``One-Shot'' approach, users treat LLMs as high-capability search engines, expecting a flawless engineering specification from a single inference. While initial outputs often appear syntactically impressive, they frequently harbor latent physical contradictions that surface only during downstream testing---at far greater cost than the original generation. CAAF accepts initial output variance and treats the first generation as a ``coarse expansion'' of possibilities, then systematically prunes this draft via Structured Semantic Gradients until it aligns with physical reality. The cost of convergence is paid in cheap inference tokens, not in expensive engineering labor.

\subsection{The Economics of Certainty: Redefining Total Cost of Ownership}

We formalize the \textbf{Total Cost of Ownership (TCO)} for AI-generated engineering artifacts:

\begin{equation}
\text{TCO} = (N_{\text{loops}} \times C_{\text{inf}}) + (P_{\text{fail}} \times C_{\text{cat}}) + C_{\text{HITL}}
\end{equation}

\noindent where $N_{\text{loops}}$ = internal inference cycles, $C_{\text{inf}}$ = per-cycle inference cost, $P_{\text{fail}}$ = probability of an undetected defect reaching production, $C_{\text{cat}}$ = catastrophic failure cost (recall or redesign), and $C_{\text{HITL}}$ = human debugging labor cost.

Monolithic deployments minimize $N_{\text{loops}}$ (fast single-shot generation) but accept a non-zero $P_{\text{fail}}$ and incur substantial $C_{\text{HITL}}$ as engineers diagnose ``compliant hallucinations.'' CAAF executes a \textbf{Compute-for-Risk Arbitrage}: deliberately inflating $N_{\text{loops}}$ through relentless UAI validation drives $P_{\text{fail}}$ toward zero on all modeled constraints---empirically observed at 100\% detection across both reported benchmarks---while reducing $C_{\text{HITL}}$ from ``hunting for hidden paradoxes'' to ``executive decision authorization.''

\textbf{Empirical TCO Analysis (L3 AD Paradox Benchmark):}
API costs were directly measured across $n{=}30$ trials per condition. Enterprise pricing (illustrative reference rates): GPT-4o at \$2.50/\$10.00 per 1M tokens (input/output); GPT-4o-mini at \$0.15/\$0.60 per 1M tokens.

\begin{table}[htbp]
\centering
\caption{Empirical TCO Comparison. Dashes in the last row indicate the metric is undefined: no correct artifacts were produced, so ``cost per correct artifact'' is not estimable from the measured data (it would be $\infty$ in the limit of any finite API spend divided by zero successes).}
\label{tab:tco}
\adjustbox{max width=\textwidth}{%
\begin{tabular}{lccc}
\toprule
Metric & Mono GPT-4o & Mono GPT-4o-mini & CAAF-all-mini \\
\midrule
API cost per run (measured) & \$0.0145 & \$0.0006 & \$0.0027 \\
Paradox detection rate & 0\% & 0\% & \textbf{100\%} \\
Artifact physically verified? & No & No & \textbf{Yes} \\
Estimated HITL cost (per failure) & \$50.00 & \$50.00 & \$5.00 \\
Effective cost per \textit{correct} artifact & ---\,\textsuperscript{\textdagger} & ---\,\textsuperscript{\textdagger} & \textbf{\$0.0027} \\
\bottomrule
\end{tabular}}

\smallskip
{\footnotesize \textsuperscript{\textdagger}Not estimable: 0\% paradox detection $\Rightarrow$ zero correct artifacts in the denominator.}
\end{table}

CAAF-all-mini costs 4.5$\times$ more per run than monolithic GPT-4o-mini (\$0.0027 vs.\ \$0.0006), yet it is the only condition that produces a valid artifact. When cost is measured per \textit{correct outcome} rather than per \textit{inference call}, monolithic deployments have no estimable cost-per-correct-artifact for this class of irreconcilable paradox---zero successes in the denominator. For deployments using on-premises open-weight models, per-run inference cost approaches near-zero, retaining reliability comparable to our reference configuration on the two reported benchmarks (empirically validated with Cohere Command-R7B and Gemma-3-12B-IT; see Finding~5 and Table~\ref{tab:open_weight_replication}) while eliminating the API cost differential entirely.

\textit{HITL cost basis: \$50/incident $=$ $\sim$20 min of senior systems engineer time to diagnose a ``compliant hallucination''; \$5/incident $=$ $\sim$2 min of executive review of a pre-computed boundary report. These are conservative engineering-labor estimates for illustration; actual values are domain- and organization-specific.}

\textbf{Scaling Behavior:}
For tasks with $S$ independent constraints, monolithic joint success probability decays as $P_{\text{success}} = p^S$ (e.g., $p{=}0.85$ per constraint). CAAF's constraint-wise decomposition breaks this joint-failure dependency, providing approximately linear cost scaling. Table~\ref{tab:scaling} projects the theoretical TCO differential as constraint complexity increases:

\begin{table}[htbp]
\centering
\caption{Theoretical TCO Scaling (Illustrative Projection)}
\label{tab:scaling}
\adjustbox{max width=\textwidth}{%
\begin{tabular}{cccc}
\toprule
Complexity (\# Constraints) & Monolithic TCO (est.) & CAAF TCO (est.) & Cost Arbitrage \\
\midrule
1 & \$8.84 & \$5.02 & 1.8$\times$ \\
5 & \$62.73 & \$5.10 & 12.3$\times$ \\
10 & \$204.06 & \$5.21 & 39.2$\times$ \\
\bottomrule
\end{tabular}}
\end{table}

\noindent\textit{These projections apply the empirical per-run API costs and the HITL cost estimates above across varying constraint counts. They are theoretical projections intended to illustrate scaling behavior, not empirical measurements at each complexity level. External UAI simulation latency (e.g., FEA solvers) would shift the bottleneck from human cognition to machine compute, reinforcing the economic thesis.}

\begin{figure}[htbp]
\centering
\includegraphics[width=0.75\textwidth]{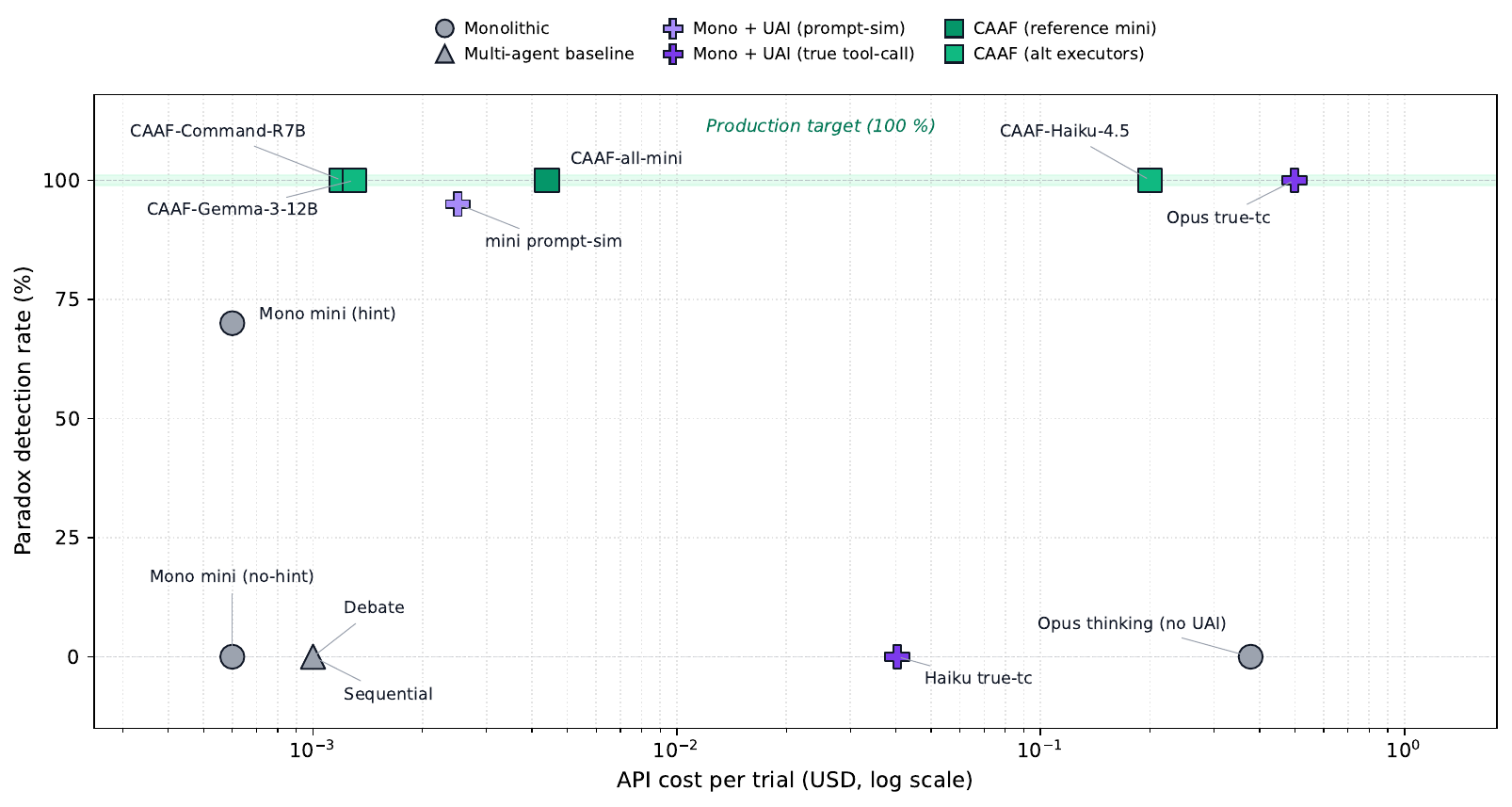}
\caption{Cost vs.\ reliability on the Pharma benchmark ($n{=}20$ per cell except where otherwise stated). CAAF configurations (green squares) cluster on the 100\% detection ceiling at \$0.0012--\$0.20 per trial. Mono+UAI true tool-call splits by reasoning tier: Opus~4 thinking reaches 100\% at \$0.499/trial; Haiku~4.5 (no thinking) reaches 0\% at \$0.0404/trial despite the same tool interface. All non-UAI baselines (monolithic, debate, sequential) sit on or near the 0\% floor. The log-scale cost axis makes the $\sim$114$\times$ gap between CAAF-all-mini and Mono+UAI-Opus visible at identical reliability.}
\label{fig:cost-vs-reliability}
\end{figure}

\subsection{Pharma Cost Comparison at 100\% Reliability}
\label{sec:pharma-cost}

Table~\ref{tab:cost-pharma} consolidates the per-trial API economics of every configuration that achieves 100\% paradox detection on the 7-constraint Pharma benchmark, sorted by cost. The frontier-reasoning Mono+UAI true-tool-call path matches CAAF's reliability but costs 114$\times$ more per trial than CAAF-all-mini and 415$\times$ more than CAAF-Command-R7B---driven by the per-trial mean of 3.15 sequential tool-call iterations on Opus~4 thinking.

\begin{table}[htbp]
\centering
\caption{Pharma per-trial cost at 100\% paradox detection ($n{=}20$ each, no-hint).}
\label{tab:cost-pharma}
\small
\adjustbox{max width=\textwidth}{%
\begin{tabular}{llccc}
\toprule
System & Executor & Iters & \$/trial & vs.\ CAAF-mini \\
\midrule
CAAF                              & Cohere Command-R7B (open)           & ---  & \$0.0012 & 0.27$\times$ \\
CAAF                              & Gemma-3-12B-IT (open)               & ---  & \$0.0013 & 0.30$\times$ \\
\textbf{CAAF (reference)}         & GPT-4o-mini                         & ---  & \$0.0044 & 1$\times$ \\
CAAF                              & Claude Haiku 4.5                    & ---  & \$0.20   & 46$\times$ \\
Mono $+$ UAI (true tool-call)     & Claude Opus~4 thinking, effort=high & 3.15 & \$0.499  & 114$\times$ \\
\bottomrule
\end{tabular}}
\end{table}

\textbf{Mono+UAI is not a cheap alternative at commodity-model cost.} The obvious steelman---pairing a cheap executor with a real tool-call UAI interface instead of running the CAAF pipeline---is empirically refuted by Cond.~7: Claude Haiku~4.5 (no extended thinking), granted the same true tool-call UAI access, exhausts \texttt{max\_iters} on 20/20 Pharma trials at \$0.0404/trial and reaches 0\% paradox detection (Finding 8, Table~\ref{tab:pharma_results}). The cheapest configuration that both (a)~uses a commodity model and (b)~reaches 100\% is CAAF-all-mini at \$0.0044/trial, which is $\sim$9$\times$ cheaper than commodity Mono+UAI \emph{and} $\infty\times$ more reliable. This is the empirical content of Finding~10: deterministic UAI is necessary but not sufficient; at commodity-model cost, reliability requires the Structured Semantic Gradient with State Locking (Pillar~3) to prevent the oscillation trap. The architectural value CAAF adds over Mono+UAI is therefore both (i)~closing the 0\% $\to$ 100\% reliability gap at commodity cost and (ii)~the $\sim$114$\times$ cost reduction vs.\ the frontier-reasoning Mono+UAI path.

\subsection{The AI as Collaborative Decision Arbiter}

Post-mortem analyses of catastrophic engineering failures---including the space shuttle Challenger O-ring incident and the Boeing 737 MAX Maneuvering Characteristics Augmentation System (MCAS) design defects---have identified systemic organizational factors as significant contributors alongside technical causes \citep{vaughan1996challenger}. In these cases, engineers possessed relevant technical knowledge but were unable to prevent the erosion of safety margins under schedule and managerial pressure. The common pattern is not ignorance of risk but institutional failure to preserve engineering red lines when they conflict with delivery timelines.

CAAF structurally addresses this behavioral vulnerability. Consider a scenario where a project manager requests that a software team ``bypass the safety warning'' to meet a deployment deadline. A monolithic LLM may comply. CAAF does not, by construction: if the artifact violates the frozen Harness, the UAI emits \texttt{[HARD FAILURE]}. To proceed, the team must engage the Strategic Negotiation process (Section~3.5), creating a formal, timestamped decision record. CAAF is designed to convert transient project pressures into durable institutional knowledge, conditional on the integrity of the Harness Registry itself (see Section~12.2).

\subsection{A Constraint Satisfaction Layer for LLM-Generated Artifacts}

The empirical results in this paper span two domains---L3 AD requirements engineering (Section~4) and pharmaceutical continuous flow reactor design (Section~\ref{sec:pharma}). We argue that the problem CAAF addresses is not limited to these, and offer the framing below as a hypothesis motivated by the architecture---one whose empirical scope remains bounded by the two benchmarks reported here. We conjecture that workflows of this shape---an LLM generating a complex artifact subject to formally verifiable constraints---share the same structural risk surface demonstrated in Sections~4--8: sycophantic compliance, context rot, stochastic oscillation, and silent constraint override; broader empirical validation of this conjecture remains future work. The pattern is general: \textbf{human intent expressed in natural language $\rightarrow$ LLM-generated artifact $\rightarrow$ formal correctness criteria that must all be satisfied simultaneously $\rightarrow$ potential contradictions among criteria that must be detected, not masked}.

CAAF's domain-specific component is exclusively the Harness Registry and its UAI validators. The RAD decomposition engine, convergence control loop, State Locking mechanism, and Strategic Negotiation protocol are designed to be domain-agnostic. Replacing the YAML Harness with domain-appropriate assertions is the intended extension mechanism, and we conjecture that the convergence behavior observed on the two benchmarks carries over to any domain where constraints can be formalized as executable predicates and composed into a DAG with bounded branching factor. The UAI evaluation mechanism is identical whether the assertion computes a braking distance, a thermodynamic equilibrium, a regulatory compliance check, or a financial risk exposure---but empirical validation beyond physical-law constraints remains future work.

We identify four categories of formally verifiable constraints, each mapping to distinct industries:

\begin{itemize}
\item \textbf{Physical Laws} (kinematics, thermodynamics, material science): \textit{Bio-pharma process design}---temperature ramp stability can be formalized as thermodynamic assertions; UAI would evaluate whether proposed cooling schedules violate phase transition boundaries. \textit{Structural engineering}---load-bearing calculations can be expressed as Finite Element Analysis (FEA)-derived assertions; existing simulation tools can be integrated via MCP as UAI validators.

\item \textbf{Regulatory and Legal Rules} (ISO standards, FDA regulations, GDPR, building codes): \textit{Clinical trial protocol design}---sample size, exclusion criteria, and informed consent requirements are formalizable constraints that frequently conflict with practical constraints (budget, timeline, patient availability). CAAF's paradox detection and Strategic Negotiation are directly applicable: when ethical constraints conflict with statistical power requirements, the system surfaces the trade-off with quantified impact rather than silently relaxing either.

\item \textbf{Mathematical and Logical Rules} (type systems, formal specifications, financial models): \textit{Financial risk modeling}---regulatory capital requirements (e.g., Basel III/IV) impose simultaneous constraints on leverage ratios, liquidity coverage, and risk-weighted assets. These constraints are mathematically expressible and frequently tension against business objectives.

\item \textbf{Infrastructure and Operational SLAs} (latency, availability, cost budgets): \textit{Cloud Infrastructure as Code}---Kubernetes deployment configurations must simultaneously satisfy security policies (network isolation, pod security standards), SLAs (latency $<$ 100ms, replicas $\geq$ 3), and cost budgets. These three constraint categories are individually formalizable and routinely conflict: high availability requires more replicas, which violates cost constraints. Current tools (OPA, Terraform plan) perform post-hoc policy checks but do not provide convergence or paradox detection.
\end{itemize}

\textbf{Embodied AI and Vision-Language-Action (VLA) Models.} A particularly compelling application frontier is the emerging class of Vision-Language-Action models (e.g., RT-2 \citep{brohan2023rt2}, Octo \citep{team2024octo}) for robotics and autonomous vehicles. VLA models generate action plans from visual perception and natural language instructions, but remain fundamentally probabilistic---they provide no formal guarantee that a proposed plan satisfies safety constraints. Current safety mechanisms in embodied AI operate primarily at the \textbf{control layer} (millisecond-level): Control Barrier Functions (CBFs) enforce real-time safety boundaries, action-space clipping limits individual joint ranges, and emergency-stop monitors provide a last resort. However, at the \textbf{planning layer} (second-level)---where VLA models decide \textit{what to do}---there is no equivalent of a constraint satisfaction guarantee: if a VLA proposes a manipulation plan that simultaneously violates gripper force limits, collision boundaries, and task-completion deadlines, no existing system detects the multi-constraint conflict, converges toward a feasible plan, or identifies when the constraint set is irreconcilable.

CAAF fills this gap as a \textbf{planning-layer constraint satisfaction layer} that complements control-layer safety mechanisms. Each VLA-proposed action plan can be validated against a Harness encoding physical safety constraints (joint torque limits, collision boundaries), task constraints (completion time, precision tolerances), and human-safety constraints (approach speed, handover force). When constraints conflict---for instance, when a manipulation task requires a trajectory that is fast enough to meet a deadline but too fast for safe human proximity---CAAF's paradox detection surfaces the trade-off with quantified options rather than silently selecting a compromised plan. This positions CAAF and CBFs as complementary layers: CAAF ensures the plan is correct \textit{before execution}; CBFs ensure execution stays safe \textit{during execution}.

In each case across the domains above, the distinguishing feature of CAAF is not merely constraint checking (which existing tools already provide) but the combination of \textbf{convergence toward satisfaction}, \textbf{paradox detection when the constraint set is irreconcilable}, and \textbf{structured negotiation with quantified trade-offs}. No existing framework provides all three capabilities in a unified architecture.

We provide empirical validation for the first category (Physical Laws) in Section~\ref{sec:pharma}. Validation across the remaining categories constitutes a primary direction for future research (Section~12.6).

\textbf{Enterprise Offline Deployment.} The experimental results (Finding~5, Section~4.2, Table~\ref{tab:open_weight_replication}) empirically establish that CAAF's reliability does not depend on frontier closed-source model capability. Two open-weight families at two distinct parameter scales---Cohere Command-R7B (7B, structured-output fine-tuned) and Google Gemma-3-12B-IT (12B, general-purpose)---each achieve 100\% on both the AD PASS-path benchmark (20/20 SUCCESS) and the pharma 3-way paradox benchmark (20/20 paradox detection), exactly matching the GPT-4o-mini reference configuration. For regulated industries where data-residency constraints prohibit transmission to external APIs (automotive type approval, ISO~13485 medical devices, IEC~62443 industrial control systems \citep{iso26262}), a fully offline CAAF instance using one of these open-weight models can in principle be certified as part of the engineering toolchain without sacrificing the convergence guarantee. The Harness Registry and UAI remain deterministic regardless of model host, and the convergence behavior is architectural rather than API-dependent. This positions CAAF as a practical path to AI-augmented requirements engineering in environments currently closed to cloud-dependent LLM tooling, with the one engineering prerequisite surfaced by our replication study: the selected open-weight model must either (a)~be fine-tuned for structured / tool-use output, or (b)~be large enough ($\ge$12B) that general instruction-following is sufficient to emit schema-faithful JSON for the Orchestrator's decomposition step.

\section{Related Work}
\label{sec:related-work}

\subsection{LLM Agent Frameworks}

\textbf{Prompt Engineering} treats the LLM as a black-box stochastic optimizer, refining inputs to maximize the probability of a desirable output distribution. CAAF explicitly supersedes this approach: rather than optimizing the \textit{input} to a single inference, it applies a formal cybernetic control loop to the \textit{output} across multiple iterations.

\textbf{ReAct} \citep{yao2023react} introduced the Thought$\rightarrow$Action$\rightarrow$Observation loop. CAAF extends this paradigm by externalizing the ``Observation'' phase into a computationally isolated, deterministic Reviewer---enforcing the Strict Role Isolation that \citet{huang2024cannot} demonstrate is necessary for reliable complex reasoning.

\textbf{Orchestration and Multi-Agent Frameworks.} LangGraph \citep{langgraph2024} provides durable graph-based orchestration but admits ``Node Logic Leakage'' absent manual per-node guards. AutoGen \citep{wu2023autogen} and Multiagent Debate \citep{du2024improving} enable social-handoff actor patterns; in constraint-satisfaction settings these are susceptible to Stochastic Oscillation (Section~6) and False Consensus (Section~8). CAAF replaces qualitative social handoffs with unarguable semantic gradients mechanically mapped to UAI results. AgentForge \citep{jafari2026agentforge} represents a recent class of lightweight modular agent frameworks that compose skills as a DAG with YAML-based configuration; CAAF complements this direction by adding a deterministic external validator (UAI) and a convergence loop on top of the same compositional primitives.

\textbf{Recursive Context Decomposition.} Recursive task decomposition with isolated subcontexts has become a commodity primitive, formalized recently as Recursive Language Models \citep{zhang2025recursivelm} and shipped in production developer-facing agent SDKs. CAAF's RAD contribution is the harness coupling described in Section~3.2: each isolated subcontext receives only the harness slice declared in its scope, so isolation functions as a compliance firewall, not just a context-window economy measure.

\subsection{Declarative LLM Programming and Computational Assertions}

The closest precedent for CAAF's Pillar~2 (Harness as an Asset) is \textbf{DSPy Assertions} \citep{khattab2023assertions} within the DSPy declarative LLM programming model \citep{khattab2023dspy}. \texttt{dspy.Assert} and \texttt{dspy.Suggest} introduce \emph{computational constraints} on LLM pipeline outputs as Python predicates that drive both generation (via assertion-driven backtracking and self-refinement) and validation, with a reported $+164\%$ constraint compliance improvement over unconstrained chains.

CAAF's HaaA shares this central insight---a single executable specification driving generator and reviewer alike---and differs in two respects oriented to safety-critical engineering. First, HaaA is \emph{asset-shaped}: a versioned registry binds YAML constraint rules to external validators (simulation tools, EDA, Hardware-in-the-Loop benches, formal verifiers such as TLA+) dispatched through the UAI, under the harness lifecycle described in Section~3.2 (test-driven validation, freezing, role-based access control, Change Approval Board); the registry therefore persists across products and regulatory revisions rather than being authored per-pipeline, and the constraint surface is not limited to what fits inside a Python predicate. Second, HaaA's failure semantics are designed as safety-case evidence: each UAI verdict carries the rule identifier, the deterministic error trace, and the locked state in a form fit for the evidence-of-compliance argumentation required by safety standards such as ISO~26262, rather than for self-refinement alone.

\subsection{LLM Guardrails and Safety}

\textbf{Constitutional AI} \citep{bai2022constitutional} encodes safety principles into model behavior through training-time feedback. This \textit{internalizes} constraints into model weights. CAAF \textit{externalizes} constraints into UAI assertions: a model weight can drift or be overridden by fine-tuning; a Python assertion cannot.

\textbf{NeMo Guardrails} \citep{rebedea2023nemo} and \textbf{Guardrails.ai} \citep{guardrailsai2023} provide programmable single-shot output validation. CAAF differs by targeting \textit{physical-semantic} compliance over a topological dependency graph, with intercept \emph{plus} convergence rather than pass/fail intercept alone.

\textbf{Consequence-Aware Agentic AI (CA2I)} \citep{anbarjafari2026ca2i} is concurrent work that introduces a pre-execution outcome-assessment module for responsible LLM agents, sharing CAAF's premise that constraint violations should be caught \emph{before} the agent acts. The two are complementary: CA2I evaluates projected consequences against high-level responsibility criteria, whereas CAAF formalizes domain invariants as an executable Harness Registry and uses a deterministic UAI plus monotonic-convergence loop to drive the agent toward a verifiable artifact.

\subsection{Textual Gradients and Optimization}

A line of recent work treats \emph{textual} feedback as the central optimization signal for LLM-driven systems. \textbf{ProTeGi} \citep{pryzant2023protegi} introduced ``natural language gradients'' for prompt search via beam search. \textbf{TextGrad} \citep{yuksekgonul2024textgrad} formalized this metaphor into an end-to-end framework that backpropagates textual gradients through arbitrary LLM workflows. \textbf{Trace} \citep{cheng2024trace} (NeurIPS 2024) generalized the same idea via the Optimization with Trace Oracle (OPTO) formalism over execution traces. \textbf{GEPA} \citep{agarwal2025gepa} (ICLR 2026 Oral) demonstrates that reflective prompt evolution leveraging natural-language feedback can outperform reinforcement learning baselines requiring thousands of rollouts.

CAAF's Structured Semantic Gradient sits in the same metaphor and differs along two dimensions. First, the optimization target: TextGrad, Trace, and GEPA optimize \emph{program parameters} (prompts, sub-prompts, workflow code) across many episodes, whereas CAAF's gradient drives in-context monotonic convergence toward a verified artifact within a \emph{single} inference, with no parameter or prompt update. Second, symbolic anchoring: the \texttt{Dimension} and \texttt{Direction} components of CAAF's gradient are produced by a deterministic boolean failure from the UAI's Python assertion engine; only the \texttt{Magnitude} is LLM-inferred. The two lines compose: an offline-optimized DSPy/GEPA pipeline could be deployed inside a CAAF Executor whose runtime artifacts are then verified by the UAI.

\subsection{Neuro-Symbolic and Formal Methods}

CAAF instantiates a practical \textbf{Neuro-Symbolic} architecture \citep{schick2023toolformer}: the LLM handles unstructured language and heuristic search, while symbolic components (Python assertion engine, simulation tools, TLA+ verifiers) provide rigorous mathematical evaluation.

\textbf{SMT Solvers (Z3, CVC4).} Satisfiability-modulo-theories solvers \citep{demoura2008z3, barrett2011cvc4} are the gold standard for deciding constraint satisfaction over formal theories. The key differentiator is \textit{the input boundary}: Z3 requires the full problem to already be expressed in \texttt{SMT-LIB}, whereas CAAF accepts natural-language requirements and lifts them into executable assertions through the Orchestrator/RAD pipeline. The two are complementary: an SMT solver is a legitimate UAI implementation choice for a mixed-integer linear subproblem. What CAAF adds is (i)~the natural-language-to-assertion lift, (ii)~monotonic convergence with State Locking rather than a single yes/no verdict, and (iii)~a paradox-surfacing protocol that returns quantified trade-offs to a human operator rather than an \texttt{unsat} core.

\textbf{Soft-FSM.} \citet{liao2026softfsm} introduces an explicitly named ``monotonic progress'' mechanism: a neuro-symbolic architecture in which an external deterministic state controller enforces monotonic accumulation of Key Information Units across a long-horizon LLM inquiry, with reported gains from $<\!40\%$ baseline completeness to $>\!97\%$ on legal cross-examination tasks. CAAF's State Locking achieves the same monotonicity invariant declaratively at the constraint-dimension level (Eq.~\ref{eq:monotonic}) rather than via an up-front procedural FSM topology---more appropriate when the constraint set is declarative and the convergence shape is emergent.

\textbf{Agent-C.} \citet{agentc2025temporal} interleaves SMT solving with constrained generation to enforce temporal safety properties on LLM agents at decode time. This is a stricter formal guarantee than CAAF provides for any individual rule but operates at a different granularity: Agent-C polices the \emph{action sequence} of an executing agent, while CAAF polices the \emph{artifact content} produced by an Executor.

\textbf{Constrained / Guided Decoding.} Systems such as Guidance \citep{microsoft2023guidance} and Outlines \citep{willard2023efficient} enforce constraints at the \textit{token level} by masking the decoder's output distribution. This is effective for \textit{structural} constraints (valid JSON, valid SQL) but does not address \textit{semantic-physical} constraints: a schema-compliant JSON document can still specify $v_{t5}{=}120\,\text{km/h}$ under heavy rain. CAAF operates at a higher layer---the \textit{artifact layer}---and verifies constraints requiring arbitrary computation. The two are compositional: constrained decoding guarantees a parseable Harness-conformant artifact, after which CAAF's UAI performs the semantic validation and gradient-based revision that constrained decoding cannot express.

\subsection{Compliance-Domain Constrained Generation}

The most direct prior art to CAAF's overall positioning is \textbf{ATLAS} \citep{ma2025atlas}, a constraint-guided generation framework for structured engineering artifacts in LLM-assisted Model-Driven Engineering. ATLAS places generation inside a model-driven workflow that separates domain representation, constraint compilation, and post-generation validation, demonstrated on AUTOSAR ARXML generation with a layered combination of XSD schemas, GBNF grammar masking, SHACL graph validators, and SMT post-validation; it reports 100\% schema validity vs.\ 0\% for unconstrained prompting. \textbf{Blueprint First, Model Second} \citep{qiu2025blueprint} argues the broader thesis that workflow logic should be separated from the generative model via an expert-defined Execution Blueprint codified as source code, with the LLM invoked only as a bounded sub-task tool.

CAAF agrees with both directions and contributes two ingredients absent from either: a closed control loop with State Locking that drives monotonic convergence to a verified artifact rather than terminating after a single constrained pass (ATLAS) or a single blueprint pass (Blueprint First); and a HaaA registry whose constraints are accumulated as a long-lived organizational asset rather than authored per-blueprint or compiled from formal artifacts alone.

\subsection{Harness Engineering as an Emerging Discipline}

\textbf{Practitioner Foundations.} \citet{bockeler2026harness} established patterns for managing LLM interaction environments to improve reliability in developer productivity tools. CAAF formalizes these patterns into a rigorous framework with guaranteed convergence properties, shifting the application domain from coding assistance (Agent-as-Tool) to safety-critical engineering governance (Agent-as-Controller). \citet{rajasekaran2026harness} at Anthropic reports a Generator--Evaluator architecture for long-running application development that is structurally isomorphic to CAAF (Planner $\approx$ RAD; Generator $\approx$ Executor; Evaluator $\approx$ UAI + Reviewer). The critical difference is that their Evaluator is itself an LLM (Playwright browser automation), whereas CAAF's UAI is a deterministic assertion engine---a distinction Finding~9 demonstrates is decisive: LLM-based evaluation achieves 0\% paradox detection across 80 trials. \citet{zhang2026harness_ad} independently maps Harness Engineering onto the ISO~21448 (SOTIF) safety lifecycle in autonomous driving.

\textbf{Concurrent Academic Work.} \citet{lee2026metaharness} introduce \emph{Meta-Harness}, an outer-loop search over harness code itself; \citet{meng2026harness_survey} survey 22 harness systems and propose labeled-transition-system semantics for agent execution loops. These confirm ``Agent Harness'' as established 2026 vocabulary. CAAF's positioning differs in that the harness is treated as a long-lived domain asset (versioned, frozen under RBAC, accumulated across products) rather than synthesized per task, and the target domain is safety-critical engineering artifacts rather than developer-tool reliability.

\subsection{Comparative Summary}

Table~\ref{tab:comparison} contrasts CAAF with prior agent and constraint-checking frameworks along the four capabilities highlighted throughout this paper: deterministic constraint checking, convergence behavior, paradox detection on irreconcilable inputs, and structured negotiation of trade-offs. To our knowledge, no prior system provides all four in a unified architecture.

\begin{table}[htbp]
\centering
\caption{Comparative summary of frameworks. Dashes (---) indicate absence. CAAF is the only framework providing all four capabilities in a unified architecture.}
\label{tab:comparison}
\adjustbox{max width=\textwidth}{%
\setlength{\tabcolsep}{8pt}
\renewcommand{\arraystretch}{1.25}
\begin{tabular}{lcccc}
\toprule
\textbf{Framework} & \textbf{Constraint Check} & \textbf{Convergence} & \textbf{Paradox Detection} & \textbf{Structured Negotiation} \\
\midrule
AutoGPT / BabyAGI & --- & --- & --- & --- \\
LangGraph \citep{langgraph2024} & Manual guards & Manual trapping & --- & --- \\
AutoGen \citep{wu2023autogen} & --- & Qual.\ reflection & --- & --- \\
NeMo / Guardrails.ai & Pass/fail & --- & --- & --- \\
OPA \citep{opa2019} & Deterministic & --- & --- & --- \\
DSPy Assert \citep{khattab2023assertions} & Python predicate & Backtrack \& retry & --- & --- \\
TextGrad \citep{yuksekgonul2024textgrad} & LLM-judged & LLM gradient & --- & --- \\
GEPA \citep{agarwal2025gepa} & LLM-judged & Reflective evolution & --- & --- \\
Z3 / SMT \citep{demoura2008z3} & Formal (SMT-LIB in) & Decision procedure & \texttt{unsat} core & --- \\
Outlines / Guidance & Token-level (schema) & --- & --- & --- \\
Soft-FSM \citep{liao2026softfsm} & Determ.\ controller & Monotonic FSM transitions & --- & --- \\
ATLAS \citep{ma2025atlas} & XSD + SHACL + SMT & Single-pass + post-validate & --- & --- \\
\midrule
\textbf{CAAF (ours)} & \textbf{Deterministic (UAI)} & \textbf{State Lock + Boundary Grad.} & \textbf{Topological RCA} & \textbf{Quantified Trade-offs + HITL} \\
\bottomrule
\end{tabular}}
\end{table}

\section{Discussion}

\paragraph{Provenance of the Three Pillars.} The mechanisms underlying CAAF's three pillars---computational assertions, structured textual feedback, monotonic state preservation, and recursive decomposition---each have independent prior art (see Section~\ref{sec:related-work}). Our contribution is the integration: a single coherent framework that couples these mechanisms with a long-lived, versioned domain harness library targeted at safety-critical engineering. This positioning makes CAAF incremental at the mechanism level and novel at the system level: removing any one pillar from the architecture is feasible (DSPy Assertions, Soft-FSM, or ATLAS each subsume one), but no prior system, to our knowledge, exposes the full set as a single asset-shaped framework with the harness lifecycle governance described in Section~3.2.

\paragraph{On Multi-Agent Decomposition.} \citet{cognition2024dontmulti} argues against multi-agent decomposition on the grounds that isolated subagents lose the implicit shared state required for coherent behavior. The case is well taken in \emph{open-ended} application development. In safety-critical engineering, however, shared implicit state is precisely the contamination vector we wish to eliminate: the ``compliant hallucination'' paradox of Section~2.1 occurs when budgetary context leaks into a safety reasoning step. CAAF therefore takes the opposite normative position for the safety-critical regime---isolation is a feature, and explicit cross-node validation through the Constraint Awareness Layer takes the place of implicit shared state.

\section{Limitations and Future Work}

\subsection{Orchestrator Context Horizon}

CAAF's RAD mitigates structural hallucination via deterministic parsing of Harness metadata. However, in hyper-complex industrial systems (e.g., Very-Large-Scale Integration (VLSI) chip design, macro-scale supply chains), even the Orchestrator may eventually face context saturation when managing thousands of interdependent sub-graphs.

We propose \textbf{Recursive Sub-Graph Partitioning (RSP)} as the mitigation strategy. The master Orchestrator applies a topological clustering algorithm (e.g., Louvain or Spectral Clustering) to divide the global graph into semantically cohesive sub-domains, each managed by a dedicated Sub-Orchestrator. Global convergence is achieved via a recursive merge-and-audit step at domain interfaces. This hierarchical approach scales CAAF logic to systems of arbitrary complexity.

\subsection{Harness Fidelity Bottleneck}

CAAF's determinism guarantees are bounded by Harness completeness. For domains where physical invariants are poorly understood or mathematically unformalized (e.g., emergent market sentiment, aesthetic design preferences), the UAI cannot provide grounding signals. In such domains, CAAF's structural advantages are reduced, and the system degrades gracefully toward a structured multi-agent orchestration framework without convergence guarantees.

\subsection{Convergence Guarantees}

The monotonic non-regression property (Eq.~\ref{eq:monotonic}) holds for the \textit{set of modeled constraints} under the assumption that (a)~the Harness is internally consistent (no contradictory assertions), and (b)~UAI assertion evaluations are deterministic. These assumptions are guaranteed by the Harness Lifecycle process (Section~3.2). What we do \emph{not} claim is finite-time convergence in the formal dynamical-systems sense: termination is enforced by a bounded \texttt{max\_iters} retry budget, not by a contraction-mapping proof. The mathematical conditions under which $V_t \to \mathcal{C}$ within a bounded number of iterations (e.g., constraint independence, bounded solution space, step-size admissibility) remain an area for formal analysis. Recent work by \citet{anbarjafari2025singularity} develops conditions and bounds for recursive-improvement convergence at a more abstract level; importing such tools to characterize the per-iteration contraction behavior of CAAF's Reviewer$\rightarrow$UAI loop is a promising direction.

\subsection{Computational Overhead and Latency}

The CAAF-all-mini configuration incurs a 4.5$\times$ API cost premium relative to a monolithic GPT-4o-mini call (\$0.0027 vs.\ \$0.0006 per run), due to multi-stage orchestration overhead. Notably, CAAF-all-mini is 5.4$\times$ \textit{cheaper} than a monolithic GPT-4o call (\$0.0027 vs.\ \$0.0145)---the only configuration that produces a physically verified artifact---making cost a non-objection for the primary comparison. Latency overhead from serialized node execution remains a practical concern for time-sensitive workflows. Future iterations will explore \textbf{Speculative Verification}, where low-confidence Executors pre-compute multiple candidate branches in parallel to reduce serialization overhead.

\subsection{Two-Domain Validation Scope}

The empirical claims of this paper rest on two benchmark domains: L3 autonomous driving requirements engineering (Section~4) and pharmaceutical continuous flow reactor design (Section~\ref{sec:pharma}). The two benchmarks are \textbf{structurally complementary}: the AD paradox involves 2~scalar constraints with a linear gap, while the pharma paradox involves 7~constraints with nonlinear (Arrhenius-exponential) interactions and a 3-way minimal unsatisfiable subset. This structural diversity strengthens the domain-agnosticism claim. Nevertheless, both benchmarks have analytically transparent ground truth. Future work will validate CAAF on domains where ground truth is less analytically transparent (e.g., financial risk modeling, supply chain optimization) and on constraint systems with even higher dimensionality (15+ constraints).

\subsection{Broader Multi-Domain Empirical Validation}

Extend the empirical evaluation to additional constraint-paradox benchmarks beyond the two domains validated here (e.g., financial regulatory compliance, supply chain logistics, energy grid optimization). Each benchmark requires a formalized Harness with verifiable ground truth. Validating CAAF on constraint systems where single-pass reasoning degrades (15+ constraints, multi-page specifications) would substantiate the hypothesis that RAD's contribution becomes critical at scale---a prediction arising from the Mono+UAI ablation results (Section~\ref{sec:pharma}).

\subsection{Automated Multi-Objective Constraint Negotiation}

Currently, constraint relaxation requires human-in-the-loop authorization. Future iterations will formalize the negotiation phase by extending the Harness schema with hierarchical variable definitions (\texttt{fixed} vs.\ \texttt{negotiable}) and \texttt{cost} attributes encoding multi-dimensional penalty functions (time, cost, safety margin, brand reputation). When the UAI detects an unresolvable paradox, a symbolic solver (e.g., SymPy-based algebraic engine) will automatically formulate the resolution as a \textbf{Multi-Objective Constrained Optimization} problem, presenting human operators with mathematically verified, cost-optimal relaxation paths.

\section{Conclusion}

As AI transitions from creative exploration to industrial governance, the primary engineering challenge shifts from maximizing raw intelligence to enforcing deterministic control. CAAF addresses this challenge by supplanting fragile probabilistic self-correction with a rigorous cybernetic feedback loop anchored in physical reality. By formalizing domain knowledge into an executable Harness and utilizing RAD to maintain context fidelity, CAAF provides a pathway toward substantially higher-reliability AI-driven engineering systems within the scope of a well-maintained Harness.

The three-tier Mono+UAI ablation (Section~\ref{sec:pharma}, Findings~8--10) reveals a more nuanced story than a single-primitive account. The deterministic UAI assertion engine is necessary but not sufficient: prompt-simulated UAI on commodity GPT-4o-mini reaches 95\%, true tool-call UAI on frontier Opus~4 thinking reaches 100\%, but true tool-call UAI on commodity-tier Haiku 4.5 reaches \textbf{0\%}---every trial oscillates between the conversion and impurity boundaries without declaring paradox. Frontier reasoning is a viable substitute for CAAF's scaffolding, but at $\sim$114$\times$ the per-trial cost of a CAAF-orchestrated commodity model. The architectural value of CAAF is therefore twofold---\emph{enforcement} (UAI invocation is an orchestrator-imposed primitive rather than an agent-discretionary tool call) and \emph{oscillation control} (Structured Semantic Gradients with State Locking prevent the stochastic-oscillation trap that destroys commodity Mono+UAI)---and the resulting cost structure makes deterministic constraint satisfaction practical at industrial scale, including in fully offline regulated environments. The three pillars remain complementary, but the reliability-load-bearing ones at commodity-model cost are UAI (Pillar~2) and Structured Semantic Gradients with State Locking (Pillar~3). RAD (Pillar~1) contributes context fidelity at scale; we hypothesize its independent reliability contribution becomes measurable at higher constraint counts (15+) where single-pass reasoning degrades---a key axis for future validation.

More broadly, we propose CAAF as a candidate architectural pattern---a \textbf{constraint satisfaction layer for LLM-generated artifacts}---intended to apply wherever the output of an LLM must simultaneously satisfy formally verifiable constraints. Just as type systems catch programming errors at compile time before code reaches production, CAAF catches constraint violations at generation time before engineering artifacts enter downstream processes. The domain-specific component is exclusively the Harness Registry; the convergence mechanism, paradox detection, and structured negotiation are designed to be domain-agnostic. On the two reported benchmarks the architectural behavior is identical: either the artifact satisfies all constraints, or the system formally reports which constraints are irreconcilable and presents quantified resolution options for human authorization. Extending this validation to regulatory, financial, and infrastructure-SLA constraint families is the primary direction for future work (Section~12.6). Subject to that broader validation, we view CAAF not as a domain-specific tool but as a prospective reliability layer for the emerging class of LLM-augmented engineering workflows in constraint-governed industries.

\appendix
\renewcommand{\thefigure}{\arabic{figure}}
\renewcommand{\thetable}{\arabic{table}}

\section{Domain-Specific Harness Assets (AD Degradation)}

The harness below is the exact production file used in the experiment (see \path{harness/data/ad_degradation.yaml} in the repository). Physics constants (e.g., \texttt{vehicle\_speed\_kmph\_t0}) are injected by the UAI at assertion time and are never sent to the LLM, preventing the model from reasoning around physical boundaries.

\begin{lstlisting}[language={},caption={AD Degradation Harness (YAML)}]
# OpenCAAF Domain Constraints: L3 Autonomous Driving Degradation
rules:
  - id: REAR_COLLISION_PREVENTION_DECELERATION
    description: "Rear Safety Constraint: To prevent rear-end collisions
      during the takeover window, the velocity drop must not exceed
      the deceleration limit."
    target_field: "vehicle_speed_kmph_t5"
    condition: "(vehicle_speed_kmph_t0 - vehicle_speed_kmph_t5) /
      (transition_window_seconds * m_per_sec_to_km_per_h_factor)
      <= max_deceleration_limit"
    assertion: "(input.get('vehicle_speed_kmph_t0')
      - input.get('vehicle_speed_kmph_t5')) /
      (input.get('transition_window_seconds')
      * input.get('m_per_sec_to_km_per_h_factor'))
      <= input.get('max_deceleration_limit')"
    severity: "CRITICAL"

  - id: FORWARD_COLLISION_PREVENTION_PERCEPTION
    description: "Forward Safety Paradox: To prevent a blind forward
      collision, the physical braking distance at the target speed
      MUST be strictly less than the perception range."
    target_field: "vehicle_speed_kmph_t5"
    condition: "((vehicle_speed_kmph_t5 / m_per_sec_to_km_per_h_factor)
      ** 2) / (2 * road_friction_mu * g) < perception_range_limit"
    assertion: "((input.get('vehicle_speed_kmph_t5')
      / input.get('m_per_sec_to_km_per_h_factor')) ** 2)
      / (2 * input.get('road_friction_mu') * input.get('g'))
      < input.get('perception_range_limit')"
    severity: "FATAL"
\end{lstlisting}

\section{Multi-Agent Convergence Trace (AD Paradox)}

The trace below shows the CAAF pipeline reaching \texttt{FAILED\_PARADOX}. All roles (Orchestrator, Executor, Reviewer) use \texttt{gpt-4o-mini}, consistent with the all-commodity-model configuration validated in the batch experiment.

\begin{lstlisting}[language={},caption={CAAF Convergence Trace (JSON)}]
{
  "step": "Decomposition",
  "role": "strategy_engine",
  "model": "gpt-4o-mini",
  "nodes": {
    "node_1": {
      "id": "node_1",
      "parent_id": null,
      "description": "Define safe-state speed limit from perception data",
      "context_keys": [],
      "expected_schema": { "perception_range_m": "float" }
    },
    "node_2": {
      "id": "node_2",
      "parent_id": "node_1",
      "description": "Calculate stopping distance at target speed",
      "context_keys": ["perception_range_m"],
      "expected_schema": { "vehicle_speed_kmph_t5": "int" }
    }
  },
  "review": {
    "id": "FORWARD_COLLISION_PREVENTION_PERCEPTION",
    "status": "FAIL",
    "error": "Stopping distance (82m) > perception limit (30m)"
  }
}
\end{lstlisting}

\section{Formal Deadlock Evidence Package}

\begin{lstlisting}[language={},caption={Formal Deadlock Evidence Package}]
[SYSTEM DEADLOCK] Formal Paradox Report
Status: FAILED_PARADOX
Domain: L3 Autonomous Driving Degradation

--- Conflict Summary ---
The system has encountered a mathematical deadlock between two
non-negotiable physical red lines:
  1. REAR_COLLISION: To prevent highway rear-end collision,
     target speed must be >= 84 km/h.
  2. FORWARD_COLLISION: To stop within 30m perception range,
     target speed must be <= 55 km/h.

--- Evidence ---
  - Attempting speed = 84 km/h: Forward Safety FAIL
    (70m stopping distance > 30m vision)
  - Attempting speed = 55 km/h: Rear Safety FAIL
    (decel = 3.6 m/s^2 > 2.0 m/s^2 limit)

--- Resolution ---
Human strategic authorization required.
See Strategic Resolution Menu.
\end{lstlisting}

\section{Reproducibility and Code Availability}

All experiments described in this paper are fully reproducible. The complete codebase, including the OpenCAAF framework implementation, harness definitions, experiment scripts, and raw result logs, is available at:

\textbf{Repository} (Apache-2.0; see project README for setup instructions): \url{https://github.com/TianbaoZhang001/OpenCAAF}

\textbf{How to run.} Each benchmark is a self-contained Python module invoked from the project root. Setup requires an OpenAI API key in \texttt{.env} (\texttt{OPENAI\_API\_KEY}) and \texttt{pip install -r requirements.txt}. The sample size for any benchmark can be overridden via the \texttt{N\_TRIALS} environment variable (default $n{=}20$, except \texttt{benchmark\_full\_experiment.py} which defaults to $n{=}30$). Example:
\begin{lstlisting}[language=bash]
N_TRIALS=20 python -m OpenCAAF.demos.benchmark_ad_pass_path
\end{lstlisting}
\begingroup\sloppy
Each run writes a timestamped directory under \path{OpenCAAF/demos/logs/} containing \path{results.json} (aggregate metrics), \path{*_runs.jsonl} (per-trial records), and, for CAAF runs, a \path{*_traces/run_NN/} subtree of per-iteration strategy plans, expert outputs, UAI feedback, and final artifacts.

\textbf{Key experiment files:}
\begin{itemize}
\item \path{OpenCAAF/demos/benchmark_full_experiment.py} --- 7-condition AD batch experiment ($n{=}30$, Table~\ref{tab:results})
\item \path{OpenCAAF/demos/benchmark_ad_pass_path.py} --- AD PASS-path convergence on satisfiable variant ($n{=}20$, Table~\ref{tab:ad-pass}, Figure~\ref{fig:ad-solution-zone})
\item \path{OpenCAAF/demos/benchmark_pharma_reactor.py} --- Pharmaceutical Flow Reactor benchmark ($n{=}20$, Table~\ref{tab:pharma_results})
\item \path{OpenCAAF/demos/benchmark_debate_baseline.py} --- Multi-agent debate baseline ($n{=}20$, Table~\ref{tab:baselines})
\item \path{OpenCAAF/demos/benchmark_sequential_baseline.py} --- Sequential checker baseline ($n{=}20$, Table~\ref{tab:baselines})
\item \path{OpenCAAF/demos/benchmark_context_rot_v2.py} --- Context Rot benchmark ($n{=}20$, Table~\ref{tab:context_rot})
\item \path{OpenCAAF/demos/benchmark_oscillation_v2.py} --- Oscillation benchmark ($n{=}20$, Table~\ref{tab:oscillation})
\item \path{OpenCAAF/demos/benchmark_pharma_pass_path.py} --- Pharma PASS-path variant ($\tau_{\max}{=}180$\,s) --- companion to the PASS-path study, retained as a harder-setting benchmark for stronger executor models (not used in the reported results)
\item \path{OpenCAAF/harness/data/ad_degradation.yaml} --- AD Harness Registry (Appendix~A)
\item \path{OpenCAAF/harness/data/ad_degradation_pass.yaml} --- AD Harness Registry (PASS-path variant, §\ref{sec:ad-pass-path})
\item \path{OpenCAAF/harness/data/pharma_flow_reactor.yaml} --- Pharmaceutical Flow Reactor Harness Registry
\item \path{OpenCAAF/harness/data/pharma_flow_reactor_pass.yaml} --- Pharmaceutical Flow Reactor Harness Registry (PASS-path variant, $\tau_{\max}{=}180$\,s)
\item \path{OpenCAAF/demos/generate_paper_figures_v2.py} --- regenerates paper figures (AD and Pharma) from logged results, including Figure~\ref{fig:ad-solution-zone}
\item \path{OpenCAAF/demos/aggregate_v2_results.py} --- scans \path{OpenCAAF/demos/logs/} and emits the unified results matrix and per-trial cost roll-up consumed by the figure script
\end{itemize}
\endgroup

\textbf{Raw result logs} are archived in \path{OpenCAAF/demos/logs/} and include per-run API traces, CAAF decision trees, and UAI assertion outputs.

The total API cost of all experiments reported in this paper---including the primary AD experiment (\$1.27), pharmaceutical reactor benchmark (\$0.14), debate and sequential baselines (\$0.34), context rot / oscillation benchmarks, and the AD PASS-path study (\$0.36)---was under \textbf{\$2.20 USD} total, demonstrating the reproducibility economics of the CAAF approach.

\bibliographystyle{plainnat}
\bibliography{references}

@inproceedings{vaswani2017attention,
  title     = {Attention Is All You Need},
  author    = {Vaswani, Ashish and Shazeer, Noam and Parmar, Niki and Uszkoreit, Jakob and Jones, Llion and Gomez, Aidan N. and Kaiser, {\L}ukasz and Polosukhin, Illia},
  booktitle = {Advances in Neural Information Processing Systems},
  volume    = {30},
  year      = {2017},
  publisher = {Curran Associates, Inc.},
  url       = {https://arxiv.org/abs/1706.03762}
}

@article{kadavath2022language,
  title   = {Language Models (Mostly) Know What They Know},
  author  = {Kadavath, Saurav and Conerly, Tom and Askell, Amanda and Henighan, Tom and Drain, Dawn and Perez, Ethan and others},
  journal = {arXiv preprint arXiv:2207.05221},
  year    = {2022},
  url     = {https://arxiv.org/abs/2207.05221}
}

@book{vaughan1996challenger,
  title     = {The Challenger Launch Decision: Risky Technology, Culture, and Deviance at NASA},
  author    = {Vaughan, Diane},
  year      = {1996},
  publisher = {University of Chicago Press}
}

@inproceedings{huang2024cannot,
  title     = {Large Language Models Cannot Self-Correct Reasoning Yet},
  author    = {Huang, Jie and Chen, Xinyun and Mishra, Swaroop and Zheng, Huaixiu Steven and Yu, Adams Wei and Song, Xinying and Zhou, Denny},
  booktitle = {The Twelfth International Conference on Learning Representations},
  year      = {2024},
  url       = {https://arxiv.org/abs/2310.01848}
}

@article{liu2024lost,
  title   = {Lost in the Middle: How Language Models Use Long Contexts},
  author  = {Liu, Nelson F. and Lin, Kevin and Hewitt, John and Paranjape, Ashwin and Bevilacqua, Michele and Petroni, Fabio and Liang, Percy},
  journal = {Transactions of the Association for Computational Linguistics},
  volume  = {12},
  pages   = {157--173},
  year    = {2024},
  publisher = {MIT Press},
  url     = {https://arxiv.org/abs/2307.03172}
}

@inproceedings{zheng2024judging,
  title     = {Judging {LLM}-as-a-Judge with {MT-Bench} and Chatbot Arena},
  author    = {Zheng, Lianmin and Chiang, Wei-Lin and Sheng, Ying and Zhuang, Siyuan and Wu, Zhanghao and Zhuang, Yonghao and Lin, Zi and Li, Zhuohan and Li, Dacheng and Xing, Eric P. and Zhang, Hao and Gonzalez, Joseph E. and Stoica, Ion},
  booktitle = {Advances in Neural Information Processing Systems},
  volume    = {36},
  year      = {2024},
  url       = {https://arxiv.org/abs/2306.05685}
}

@inproceedings{wei2022chain,
  title     = {Chain-of-Thought Prompting Elicits Reasoning in Large Language Models},
  author    = {Wei, Jason and Wang, Xuezhi and Schuurmans, Dale and Bosma, Maarten and Ichter, Brian and Xia, Fei and Chi, Ed H. and Le, Quoc V. and Zhou, Denny},
  booktitle = {Advances in Neural Information Processing Systems},
  volume    = {35},
  year      = {2022},
  url       = {https://arxiv.org/abs/2201.11903}
}

@inproceedings{yao2023react,
  title     = {{ReAct}: Synergizing Reasoning and Acting in Language Models},
  author    = {Yao, Shunyu and Zhao, Jeffrey and Yu, Dian and Du, Nan and Shafran, Irina and Narasimhan, Karthik and Cao, Yuan},
  booktitle = {The Eleventh International Conference on Learning Representations},
  year      = {2023},
  url       = {https://arxiv.org/abs/2210.03629}
}

@article{wu2023autogen,
  title   = {{AutoGen}: Enabling Next-Generation {LLM} Applications via Multi-Agent Conversation},
  author  = {Wu, Qingyun and Bansal, Gagan and Zhang, Jieyu and Wu, Yiran and Zhang, Shaokun and Zhu, Erkang and Li, Beibin and Jiang, Li and Zhang, Xiaoyun and Wang, Chi},
  journal = {arXiv preprint arXiv:2308.08155},
  year    = {2023},
  url     = {https://arxiv.org/abs/2308.08155}
}

@misc{langgraph2024,
  title  = {{LangGraph}: Build Stateful, Multi-Actor Applications with {LLM}s},
  author = {{LangChain, Inc.}},
  year   = {2024},
  url    = {https://github.com/langchain-ai/langgraph},
  note   = {Open-source library for building stateful multi-agent LLM applications}
}

@inproceedings{du2024improving,
  title     = {Improving Factuality and Reasoning in Language Models through Multiagent Debate},
  author    = {Du, Yilun and Li, Shuang and Torralba, Antonio and Tenenbaum, Joshua B. and Mordatch, Igor},
  booktitle = {Proceedings of the 41st International Conference on Machine Learning},
  series    = {Proceedings of Machine Learning Research},
  volume    = {235},
  year      = {2024},
  publisher = {PMLR},
  url       = {https://arxiv.org/abs/2305.14325}
}

@inproceedings{rebedea2023nemo,
  title     = {{NeMo Guardrails}: A Toolkit for Controllable and Safe {LLM} Applications with Programmable Rails},
  author    = {Rebedea, Traian and Dinu, Razvan and Sreedhar, Makesh Narsimhan and Parisien, Christopher and Cohen, Jonathan},
  booktitle = {Proceedings of the 2023 Conference on Empirical Methods in Natural Language Processing: System Demonstrations},
  pages     = {431--445},
  year      = {2023},
  publisher = {Association for Computational Linguistics},
  url       = {https://arxiv.org/abs/2310.10501}
}

@article{bai2022constitutional,
  title   = {Constitutional {AI}: Harmlessness from {AI} Feedback},
  author  = {Bai, Yuntao and Jones, Andy and Ndousse, Kamal and Askell, Amanda and Chen, Anna and DasSarma, Nova and Drain, Dawn and Fort, Stanislav and Ganguli, Deep and Henighan, Tom and Joseph, Nicholas and Kadavath, Saurav and Kernion, Jackson and Conerly, Tom and El-Showk, Sheer and Elhage, Nelson and Hatfield-Dodds, Zac and Hernandez, Danny and Hume, Tristan and Johnston, Scott and Kravec, Shauna and Lovitt, Liane and Nanda, Neel and Olsson, Catherine and Amodei, Dario and Brown, Tom and Clark, Jack and McCandlish, Sam and Olah, Chris and Mann, Ben and Kaplan, Jared},
  journal = {arXiv preprint arXiv:2212.08073},
  year    = {2022},
  url     = {https://arxiv.org/abs/2212.08073}
}

@misc{guardrailsai2023,
  title  = {Guardrails: Adding Guardrails to Large Language Models},
  author = {{Guardrails AI}},
  year   = {2023},
  url    = {https://github.com/guardrails-ai/guardrails},
  note   = {Open-source Python library for structured and validated LLM outputs}
}

@article{yuksekgonul2024textgrad,
  title   = {{TextGrad}: Automatic ``Differentiation'' via Text},
  author  = {Yuksekgonul, Mert and Bianchi, Federico and Boen, Joseph and Liu, Sheng and Zou, James},
  journal = {arXiv preprint arXiv:2406.07496},
  year    = {2024},
  url     = {https://arxiv.org/abs/2406.07496}
}

@inproceedings{schick2023toolformer,
  title     = {Toolformer: Language Models Can Teach Themselves to Use Tools},
  author    = {Schick, Timo and Dwivedi-Yu, Jane and Dess{\`i}, Roberto and Raileanu, Roberta and Lomeli, Maria and Zettlemoyer, Luke and Cancedda, Nicola and Scialom, Thomas},
  booktitle = {Advances in Neural Information Processing Systems},
  volume    = {36},
  year      = {2023},
  publisher = {Curran Associates, Inc.},
  url       = {https://arxiv.org/abs/2302.04761}
}

@inproceedings{shen2023hugginggpt,
  title     = {{HuggingGPT}: Solving {AI} Tasks with {ChatGPT} and Its Friends in {Hugging Face}},
  author    = {Shen, Yongliang and Song, Kaitao and Tan, Xu and Li, Dongsheng and Lu, Weiming and Zhuang, Yueting},
  booktitle = {Advances in Neural Information Processing Systems},
  volume    = {36},
  year      = {2023},
  publisher = {Curran Associates, Inc.},
  url       = {https://arxiv.org/abs/2303.17580}
}

@article{meyer1992applying,
  title     = {Applying ``Design by Contract''},
  author    = {Meyer, Bertrand},
  journal   = {Computer},
  volume    = {25},
  number    = {10},
  pages     = {40--51},
  year      = {1992},
  publisher = {IEEE},
  doi       = {10.1109/2.161279}
}

@misc{opa2019,
  title  = {Open Policy Agent},
  author = {{CNCF}},
  year   = {2019},
  url    = {https://www.openpolicyagent.org},
  note   = {{CNCF} Graduated Project. Policy-as-Code engine using the Rego language.}
}

@misc{bockeler2026harness,
  author       = {B\"{o}ckeler, Birgitta},
  title        = {Harness Engineering for Coding Agent Users},
  year         = {2026},
  month        = {April},
  howpublished = {MartinFowler.com},
  note         = {Initial memo published 17 February 2026; full article 2 April 2026},
  url          = {https://martinfowler.com/articles/harness-engineering.html}
}

@misc{rajasekaran2026harness,
  author       = {Rajasekaran, Prithvi},
  title        = {Harness Design for Long-Running Application Development},
  year         = {2026},
  month        = {March},
  howpublished = {Anthropic Engineering Blog},
  url          = {https://www.anthropic.com/engineering/harness-design-long-running-apps}
}

@misc{zhang2026harness_ad,
  author       = {Zhang, Yuxin},
  title        = {Harness Engineering in Intelligent Driving: A Cross-Domain Application},
  year         = {2026},
  month        = {April},
  howpublished = {AutoZYX Blog},
  url          = {https://blog.autozyx.com/posts/harness-engineering-intelligent-driving/},
  note         = {In Chinese. Maps Harness Engineering patterns onto ISO~21448 (SOTIF) automotive safety lifecycle.}
}

@misc{anthropic2024mcp,
  title  = {Model Context Protocol},
  author = {{Anthropic}},
  year   = {2024},
  url    = {https://www.anthropic.com/news/model-context-protocol},
  note   = {Open protocol for connecting AI assistants to external data sources and tools. Specification available at \url{https://spec.modelcontextprotocol.io}}
}

@techreport{iso26262,
  title       = {{ISO} 26262: Road Vehicles -- Functional Safety},
  author      = {{International Organization for Standardization}},
  institution = {ISO},
  type        = {International Standard},
  number      = {ISO 26262},
  edition     = {2},
  year        = {2018},
  note        = {Parts 1--12. Second edition (first published 2011).}
}

@article{brohan2023rt2,
  title   = {RT-2: Vision-Language-Action Models Transfer Web Knowledge to Robotic Control},
  author  = {Brohan, Anthony and Brown, Noah and Carbajal, Justice and Chebotar, Yevgen and Chen, Xi and Choromanski, Krzysztof and others},
  journal = {arXiv preprint arXiv:2307.15818},
  year    = {2023},
  url     = {https://arxiv.org/abs/2307.15818}
}

@article{team2024octo,
  title   = {Octo: An Open-Source Generalist Robot Policy},
  author  = {{Octo Model Team} and Ghosh, Dibya and Walke, Homer and Pertsch, Karl and Black, Kevin and Mees, Oier and others},
  journal = {arXiv preprint arXiv:2405.12213},
  year    = {2024},
  url     = {https://arxiv.org/abs/2405.12213}
}

@inproceedings{demoura2008z3,
  title     = {{Z3}: An Efficient {SMT} Solver},
  author    = {de Moura, Leonardo and Bj{\o}rner, Nikolaj},
  booktitle = {Tools and Algorithms for the Construction and Analysis of Systems (TACAS)},
  series    = {Lecture Notes in Computer Science},
  volume    = {4963},
  pages     = {337--340},
  year      = {2008},
  publisher = {Springer},
  doi       = {10.1007/978-3-540-78800-3_24}
}

@inproceedings{barrett2011cvc4,
  title     = {{CVC4}},
  author    = {Barrett, Clark and Conway, Christopher L. and Deters, Morgan and Hadarean, Liana and Jovanovi{\'c}, Dejan and King, Tim and Reynolds, Andrew and Tinelli, Cesare},
  booktitle = {Computer Aided Verification (CAV)},
  series    = {Lecture Notes in Computer Science},
  volume    = {6806},
  pages     = {171--177},
  year      = {2011},
  publisher = {Springer},
  doi       = {10.1007/978-3-642-22110-1_14}
}

@article{willard2023efficient,
  title   = {Efficient Guided Generation for Large Language Models},
  author  = {Willard, Brandon T. and Louf, R{\'e}mi},
  journal = {arXiv preprint arXiv:2307.09702},
  year    = {2023},
  url     = {https://arxiv.org/abs/2307.09702},
  note    = {Reference implementation: \texttt{outlines} library.}
}

@misc{microsoft2023guidance,
  title        = {Guidance: A Guidance Language for Controlling Large Language Models},
  author       = {{Microsoft Research}},
  year         = {2023},
  howpublished = {GitHub},
  url          = {https://github.com/guidance-ai/guidance},
  note         = {Open-source toolkit for constrained (token-level) LLM generation.}
}

@article{jafari2026agentforge,
  title   = {A Lightweight Modular Framework for Constructing Autonomous Agents Driven by Large Language Models: Design, Implementation, and Applications in {AgentForge}},
  author  = {Anbar Jafari, Akbar and Ozcinar, Cagri and Anbarjafari, Gholamreza},
  journal = {arXiv preprint arXiv:2601.13383},
  year    = {2026},
  url     = {https://arxiv.org/abs/2601.13383}
}

@misc{anbarjafari2026ca2i,
  title        = {Consequence-Aware Agentic {AI}: A Pre-Execution Outcome Assessment Framework for Responsible Large Language Model Agents},
  author       = {Anbar Jafari, Akbar and Ozcinar, Cagri and Anbarjafari, Gholamreza},
  year         = {2026},
  howpublished = {PhilArchive},
  url          = {https://philarchive.org/rec/JAFCAA}
}

@article{anbarjafari2025singularity,
  title   = {A Mathematical Framework for {AI} Singularity: Conditions, Bounds, and Control of Recursive Improvement},
  author  = {Anbar Jafari, Akbar and Ozcinar, Cagri and Anbarjafari, Gholamreza},
  journal = {arXiv preprint arXiv:2511.10668},
  year    = {2025},
  url     = {https://arxiv.org/abs/2511.10668}
}

@article{khattab2023dspy,
  title   = {{DSPy}: Compiling Declarative Language Model Calls into Self-Improving Pipelines},
  author  = {Khattab, Omar and Singhvi, Arnav and Maheshwari, Paridhi and Zhang, Zhiyuan and Santhanam, Keshav and Vardhamanan, Sri and Haq, Saiful and Sharma, Ashutosh and Joshi, Thomas T. and Moazam, Hanna and Miller, Heather and Zaharia, Matei and Potts, Christopher},
  journal = {arXiv preprint arXiv:2310.03714},
  year    = {2023},
  url     = {https://arxiv.org/abs/2310.03714},
  note    = {Reference implementation: \texttt{dspy} library, $>$16k GitHub stars.}
}

@article{khattab2023assertions,
  title   = {{DSPy} Assertions: Computational Constraints for Self-Refining Language Model Pipelines},
  author  = {Singhvi, Arnav and Shetty, Manish and Tan, Shangyin and Potts, Christopher and Sen, Koushik and Zaharia, Matei and Khattab, Omar},
  journal = {arXiv preprint arXiv:2312.13382},
  year    = {2023},
  url     = {https://arxiv.org/abs/2312.13382},
  note    = {Submitted Dec 2023; revised Feb 2024. Introduces \texttt{dspy.Assert} / \texttt{dspy.Suggest} as a programming construct for computational constraints in LM pipelines, with assertion-driven backtracking and compilation strategies for more reliable systems.}
}

@inproceedings{cheng2024trace,
  title     = {Trace Is the Next {AutoDiff}: Generative Optimization with Rich Feedback, Execution Traces, and {LLMs}},
  author    = {Cheng, Ching-An and Nie, Allen and Swaminathan, Adith},
  booktitle = {Advances in Neural Information Processing Systems 37 ({NeurIPS} 2024)},
  year      = {2024},
  url       = {https://arxiv.org/abs/2406.16218},
  note      = {Frames optimization of non-differentiable workflows as ``Optimization with Trace Oracle'' (OPTO); generalizes back-propagation by capturing and propagating execution traces. Implemented as PyTorch-like Python library (\texttt{microsoft/Trace}).}
}

@article{agarwal2025gepa,
  title   = {{GEPA}: Reflective Prompt Evolution Can Outperform Reinforcement Learning},
  author  = {Agrawal, Lakshya A. and Tan, Shangyin and Soylu, Dilara and Ziems, Noah and Khare, Rishi and Opsahl-Ong, Krista and Singhvi, Arnav and Shandilya, Herumb and Ryan, Michael J. and Jiang, Meng and Potts, Christopher and Sen, Koushik and Dimakis, Alexandros G. and Stoica, Ion and Klein, Dan and Zaharia, Matei and Khattab, Omar},
  journal = {arXiv preprint arXiv:2507.19457},
  year    = {2025},
  url     = {https://arxiv.org/abs/2507.19457},
  note    = {Submitted Jul 2025; revised Feb 2026. Accepted to ICLR 2026 (Oral). Reflective prompt evolution leveraging natural-language feedback as the optimization signal; outperforms reinforcement learning baselines requiring thousands of rollouts.}
}

@inproceedings{pryzant2023protegi,
  title     = {Automatic Prompt Optimization with ``Gradient Descent'' and Beam Search},
  author    = {Pryzant, Reid and Iter, Dan and Li, Jerry and Lee, Yin Tat and Zhu, Chenguang and Zeng, Michael},
  booktitle = {Proceedings of the 2023 Conference on Empirical Methods in Natural Language Processing (EMNLP)},
  year      = {2023},
  url       = {https://arxiv.org/abs/2305.03495},
  note      = {ProTeGi: textual ``gradients'' from LLM critic guide prompt search.}
}

@article{liao2026softfsm,
  title   = {Enforcing Monotonic Progress in Legal Cross-Examination: Preventing Long-Horizon Stagnation in {LLM}-Based Inquiry},
  author  = {Liao, Hsien-Jyh},
  journal = {arXiv preprint arXiv:2602.04206},
  year    = {2026},
  url     = {https://arxiv.org/abs/2602.04206},
  note    = {Submitted Feb 2026. Identifies ``procedural stagnation'' as failure mode and proposes Soft-FSM, a neuro-symbolic architecture enforcing monotonic progress over Key Information Units via an external deterministic state controller. Reports baseline collapse below 40\% completeness vs.\ Soft-FSM consistently above 97\% on criminal homicide cases.}
}

@article{agentc2025temporal,
  title   = {Enforcing Temporal Constraints for {LLM} Agents},
  author  = {Kamath, Adharsh and Zhang, Sishen and Xu, Calvin and Ugare, Shubham and Singh, Gagandeep and Misailovic, Sasa},
  journal = {arXiv preprint arXiv:2512.23738},
  year    = {2025},
  url     = {https://arxiv.org/abs/2512.23738},
  note    = {Submitted Dec 2025; accepted to ICLR 2026 Workshop VerifAI-2. Agent-C: DSL for temporal properties $\to$ first-order logic $\to$ SMT solver interleaved with constrained generation; backtracks on non-compliant token sequences. Reports 100\% conformance / 0\% harm on retail and airline benchmarks.}
}

@article{zhang2025recursivelm,
  title   = {Recursive Language Models},
  author  = {Zhang, Alex L. and Kraska, Tim and Khattab, Omar},
  journal = {arXiv preprint arXiv:2512.24601},
  year    = {2025},
  url     = {https://arxiv.org/abs/2512.24601},
  note    = {Submitted Dec 2025; revised Jan 2026. Treats long prompts as external environment; LLM programmatically examines, decomposes, and recursively calls itself over snippets. Processes inputs up to two orders of magnitude beyond context window.}
}

@article{qiu2025blueprint,
  title   = {Blueprint First, Model Second: A Framework for Deterministic {LLM} Workflow},
  author  = {Qiu, Libin and Ye, Yuhang and Gao, Zhirong and Zou, Xide and Chen, Junfu and Gui, Ziming and Huang, Weizhi and Xue, Xiaobo and Qiu, Wenkai and Zhao, Kun},
  journal = {arXiv preprint arXiv:2508.02721},
  year    = {2025},
  url     = {https://arxiv.org/abs/2508.02721},
  note    = {Submitted Aug 2025. Source Code Agent framework: separates workflow logic from generative model via expert-defined Execution Blueprint codified as source code. LLM invoked as bounded sub-task tool; never decides workflow path. Closest prior art to CAAF's overall thesis.}
}

@article{ma2025atlas,
  title   = {{ATLAS}: A Layered Constraint-Guided Framework for Structured Artifact Generation in {LLM}-Assisted {MDE}},
  author  = {Ma, Tong and Lai, Hui and Wang, Hui and Tian, Zhenhu and Li, Chaochao and Xu, Fengjie and Fang, Ling},
  journal = {arXiv preprint arXiv:2510.25890},
  year    = {2025},
  url     = {https://arxiv.org/abs/2510.25890},
  note    = {Submitted Oct 2025. Constraint-guided generation framework for structured engineering artifacts that must satisfy schemas, domain rules, and audit requirements. Generation is placed inside a model-driven workflow separating domain representation, constraint compilation, and post-generation validation. \textbf{This paper is the most direct prior art to CAAF's overall positioning} (LLM-assisted Model-Driven Engineering, not just AUTOSAR ARXML as the demo vertical).}
}

@article{lee2026metaharness,
  title   = {{Meta-Harness}: End-to-End Optimization of Model Harnesses},
  author  = {Lee, Yoonho and Nair, Roshen and Zhang, Qizheng and Lee, Kangwook and Khattab, Omar and Finn, Chelsea},
  journal = {arXiv preprint arXiv:2603.28052},
  year    = {2026},
  url     = {https://arxiv.org/abs/2603.28052},
  note    = {Submitted Mar 2026. Searches over harness code (storage / retrieval / presentation logic) using an agentic proposer that accesses source code, scores, and execution traces through a filesystem. Argues existing optimizers under-perform because they over-aggressively compress feedback.}
}

@misc{meng2026harness_survey,
  title        = {Agent Harness for Large Language Model Agents: A Survey},
  author       = {Meng, Qianyu and Wang, Yanan and Chen, Liyi and others},
  year         = {2026},
  howpublished = {Preprints.org, manuscript 202604.0428},
  url          = {https://www.preprints.org/manuscript/202604.0428},
  note         = {Surveys 22 systems, identifies 9 cross-cutting challenges, proposes 12 research directions. Provides formal definition of agent harness with labeled-transition-system semantics distinguishing safety and liveness. Traces ``harness'' concept from software test harnesses through RL environments to modern LLM agent infrastructure. \textbf{Confirms ``Agent Harness'' is established 2026 terminology.}}
}

@misc{cognition2024dontmulti,
  title        = {Don't Build Multi-Agents},
  author       = {{Cognition AI}},
  year         = {2024},
  howpublished = {Cognition Engineering Blog},
  url          = {https://cognition.ai/blog/dont-build-multi-agents},
  note         = {Argues against multi-agent decomposition on the grounds that isolated subcontexts lose shared implicit state. CAAF responds in the Discussion section: in safety-critical domains, shared implicit state is a contamination vector and isolation is a feature.}
}

\end{document}